\documentclass[journal, 10pt, twoside, final]{IEEEtran}
\usepackage[noadjust]{cite}

\usepackage{graphicx}
\graphicspath{{./graphics/}}

\usepackage{amsmath}
\usepackage{url}

\usepackage[tablename=Tab.]{caption}

\usepackage[utf8]{inputenc}
\usepackage{nicefrac}
\usepackage{multirow}
\usepackage{subfigure}

\usepackage{flushend}

\usepackage{pifont}
\newcommand{\cmark}{\ding{51}}%
\usepackage{xcolor}

\usepackage{colortbl}
\definecolor{Petrol}{HTML}{5097AB}

\usepackage{placeins}

\usepackage{hyperref}

\usepackage{tabulary}

\usepackage[printwatermark]{xwatermark}
\usepackage{tikz}

\begin{document}

\title{Teaching Vehicles to Anticipate:\\ A Systematic Study on Probabilistic Behavior Prediction Using Large Data Sets}

\author{Florian~Wirthmüller\textsuperscript{\includegraphics[scale=0.4]{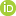}},
        Julian~Schlechtriemen\textsuperscript{\includegraphics[scale=0.4]{orcidlogo.png}},
        Jochen~Hipp\textsuperscript{\includegraphics[scale=0.4]{orcidlogo.png}} 
        and~Manfred~Reichert\textsuperscript{\includegraphics[scale=0.4]{orcidlogo.png}} %
\thanks{F. Wirthmüller, J. Schlechtriemen and J. Hipp are with Mercedes-Benz AG, Böblingen, Germany, E-Mail: \{first\_name.last\_name\}@daimler.com.}
\thanks{F. Wirthmüller and M. Reichert are with the Institute of Databases and Information Systems (DBIS), Ulm University, Ulm, Germany,\newline E-Mail: \{first\_name.last\_name\}@uni-ulm.de}
\thanks{J. Schlechtriemen is with the Institute of Realtime Learning Systems at the University of Siegen, Siegen, Germany.}
\thanks{ORCID: \href{https://orcid.org/0000-0002-9732-2561}{https://orcid.org/0000-0002-9732-2561};\newline \href{https://orcid.org/0000-0002-9130-061X}{https://orcid.org/0000-0002-9130-061X};\newline \href{https://orcid.org/0000-0002-9037-9899}{https://orcid.org/0000-0002-9037-9899};\newline \href{https://orcid.org/0000-0003-2536-4153}{https://orcid.org/0000-0003-2536-4153}}
\thanks{Manuscript received October 02, 2019; revised April 19, 2020, accepted May 19, 2020}
\thanks{\copyright~2019 IEEE. Personal use of this material is permitted. Permission from IEEE must be obtained for all other uses, in any current or future media, including reprinting/republishing this material for advertising or promotional purposes, creating new collective works, for resale or redistribution to servers or lists, or reuse of any copyrighted component of this work in other works.}}


\markboth{IEEE Transactions on Intelligent Transportation Systems (T-ITS)
}{Wirthmüller \MakeLowercase{\textit{et al.}}: Teaching Vehicles to Anticipate: A Systematic Study on Probabilistic Behavior Prediction}


\IEEEpubid{\copyright~2019 IEEE}

\maketitle

\begin{abstract}
By observing their environment as well as other traffic participants, humans are enabled to drive road vehicles safely. Vehicle passengers, however, perceive a notable difference between non-experienced and experienced drivers. In particular, they may get the impression that the latter ones anticipate what will happen in the next few moments and consider these foresights in their driving behavior. To make the driving style of automated vehicles comparable to the one of human drivers with respect to comfort and perceived safety, the aforementioned anticipation skills need to become a built-in feature of self-driving vehicles. This article provides a systematic comparison of methods and strategies to generate this intention for self-driving cars using machine learning techniques. To implement and test these algorithms we use a large data set collected over more than 30\,000\,km of highway driving and containing approximately 40\,000 real-world driving situations. We further show that it is possible to classify driving maneuvers upcoming within the next 5\,s with an Area Under the \textit{ROC} Curve (\textit{AUC}) above 0.92 for all defined maneuver classes. This enables us to predict the lateral position with a prediction horizon of 5\,s with a median lateral error of less than 0.21\,m.


\end{abstract}



\section{Introduction}

\begin{figure}[!t]
  \centering\includegraphics[width=0.8\columnwidth]{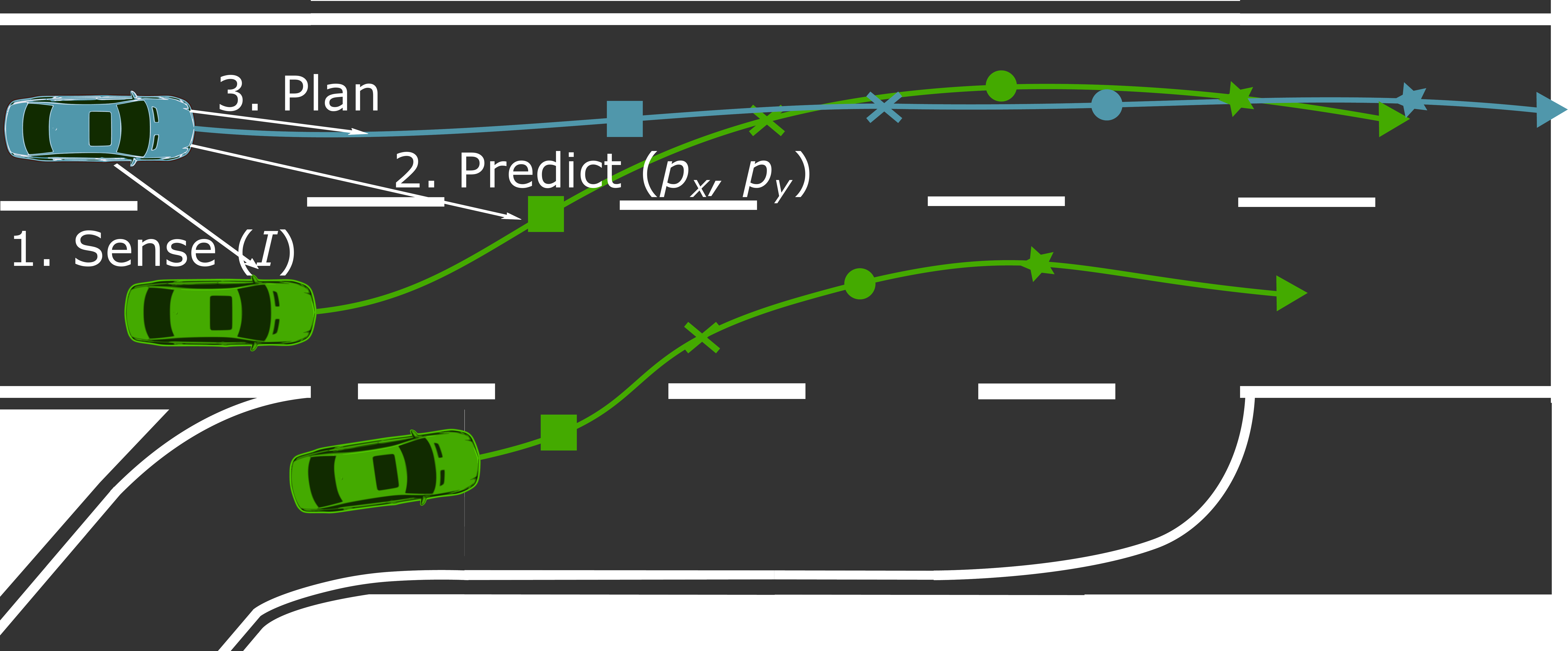}
  \caption{Long-term driving behavior predictions in the context of trajectory planning for automated driving (equal symbols denote simultaneity).}
  \label{fig:intro}
\end{figure}

\IEEEPARstart{A}{utomated driving} has the potential to radically change our mobility habits as well as the way goods are transported. To enable driving automation, several processing steps have to be executed. \autoref{fig:intro} illustrates this thought: In the first step, the current traffic scene has to be sensed and a proper representation of the environment needs to be generated. Using this information, the given traffic situation needs to be interpreted and the behavior of others has to be anticipated. Subsequently, a plan, i.\,e. a trajectory, is derived based on this knowledge. Finally, this plan is executed in the last step of this process. How long the trajectory stays viable, before it has to be re-planned, is strongly influenced by the capability of the prediction component.

As opposed to other research works dealing with techniques to interconnect vehicles through a so called car-to-car communication, we aim to solve this anticipation task locally. On one hand, it is not foreseeable when an adequate market penetration of vehicles with such techniques will be reached. On the other, a local prediction component always becomes necessary, as there are several traffic participants without communication abilities such as bicyclists. In addition, local predictions might become necessary to bypass transmission times in certain cases as  emphasized by \cite{weidl2018}. Moreover, it is reasonable to approach the topic from the perspective of highway driving, as this use case is easier to realize than others due to its clear constraints (e.\,g. structured setting, absence of pedestrians). However, for the prediction task this implies the challenge to create precise long-term predictions (2 to 5\,s) rather than short forecasts (up to 2\,s), as in highway scenarios higher velocities can be expected than in urban or rural areas. 

\IEEEpubidadjcol

\subsection{Problem Statement}\label{subsec:problem_statement}
We tackle the challenge of anticipating the behavior of other traffic participants in highway scenarios. In particular, we aim to generate information that can be processed by trajectory planning algorithms to implement an anticipatory driving style. In this context, our objective is to model future vehicle positions within a time $t$ in longitudinal $x_t$ and  lateral $y_t$ direction as spatial distributions $x_t \sim p_x$, $y_t \sim p_y$ rather than estimating single shot predictions $\hat{x}_t$ and $\hat{y}_t$ respectively. Note that these distributions are more useful for down-streamed criticality assessments as they enable us to represent several alternative hypotheses at a time with their particular frequencies. Despite the focus on highway driving, the presented methods shall be general enough to be appropriate in other environments as well.




\subsection{Problem Resolution Strategy}\label{subsec:res_strategy}
This article presents a systematic workflow for the design and evaluation of a lightweight maneuver-based model \cite{lefevre2014}, which uses standard sensor inputs to perform long-term driving behavior predictions. Methodically, we build on \cite{schlechtriemen2015will} and use a two-step Mixture of Experts (\textit{MOE}) approach. This includes a maneuver classification and a down-streamed behavior prediction. The maneuver probabilities $\{P_m\}_{\forall m \in M}$ determined by the classifier are used in the Mixture of Experts approach as gating nodes. Specifically, the probabilities control the weighting $w_m$ of the respective expert distributions $p_ {y, m}$, while calculating the overall distribution of future vehicle positions $p_y$. \autoref{eq:MOE} summarizes this procedure for the lateral direction (equivalent for $x$): 

\begin{equation}
\begin{aligned}
y_{t} & \sim p_y(\Theta_y, I, t) \\
& = \sum_{m \in M}{p_{y, m}(\theta_{y, m}| I, t) \cdot w_m(I)}\\ 
\label{eq:MOE}
\end{aligned}
\end{equation}

The set of maneuvers $M$ is defined as follows:
\begin{equation}
M = \{LCL, FLW, LCR\}
\label{eq:maneuvers}
\end{equation}

Different weighting approaches based on the maneuver probabilities are presented in \autoref{sec:traj_eval}. The expert distributions $p_{y, m}$ are modeled as Gaussian Mixture Models (\textit{GMM}s) in the combined input and output space with $K$ components according to \autoref{eq:gmm}, and are used in a Gaussian Mixture Regression manner. Hence, they are conditioned by the input features $I$ and the prediction time $t$ (cf. \autoref{eq:MOE}).

\begin{equation}
p_{y, m}(\theta_{y, m}) = \sum_{i=1}^{K}{\phi_{y, m, i} \cdot \mathcal{N}(\mu_{y, m, i}, \Sigma_{y, m, i})}
\label{eq:gmm}
\end{equation}

The parameters of the \textit{GMM}s are subsumed in $\Theta_y$:

\begin{equation}
\Theta_y = \{\theta_{y, m}\}_{\forall m \in M} = \{\phi_{y, m}, \mu_{y, m}, \Sigma_{y, m}\}_{\forall m \in M}
\end{equation}

In addition, we introduce an alternative methodology to the Mixture of Experts approach, integrating the outputs of the gating nodes into one single model. This simplifies \autoref{eq:MOE} as follows:

\begin{equation}
y_{t} \sim p_y(\theta_{y, IGMM}| I, t, P_{LCL}(I), P_{LCR}(I)) 
\label{eq:IGMM}
\end{equation}

For implementing the models, we use out-of-the-box modules from the widely used frameworks Apache Spark MLlib \cite{meng2016mllib} (classifiers) and Scikit-learn \cite{scikit-learn} (\textit{GMM}s). 

Altogether, we contribute a systematic workflow for designing and evaluating the prediction models as well as methodical extensions to known approaches. Moreover, we assess the performance of the developed modules for the two tasks of predicting (1) driving maneuvers and (2) probability distributions of future positions both separately and in combination. To evaluate the modules, we utilize a large data set comprising real-world measurements. As will be shown, our prediction models outperform established state-of-the-art approaches. \newline


The remainder of this article is organized as follows: \autoref{sec:rel_work} discusses related work on object motion prediction, emphasizing the value added by our approach. \autoref{sec:pre} introduces the data set and describes the preprocessing steps applied to it. \autoref{sec:classification} outlines the training of the considered maneuver classifiers, whereas \autoref{sec:class_eval} deals with the experimental evaluation and the performance of the classifiers. Based on these findings, \autoref{sec:traj_pred} develops different approaches for estimating probability distributions of future vehicle positions, which are then assesed in \autoref{sec:traj_eval}. Finally, \autoref{sec:conclusion} summarizes the article and gives an outlook on future work. 

\section{Related Work}\label{sec:rel_work}
Regarding the understanding and prediction of the behavior of other traffic participants in highway scenarios, various aspects were investigated in literature. Accordingly, this section is sub-divided into three parts: \autoref{subsec:rel_work_classifiers} presents approaches inferring the kind of maneuver that will be executed by a vehicle. Note that applications like collision checkers or trajectory planning algorithms cannot directly process such kind of information. Instead, probabilities of future vehicle positions or trajectories need to be predicted. Related research on this topic is presented in \autoref{subsec:rel_work_traj_pred}. Bringing together the aspects of maneuver classification and position prediction, \autoref{subsec:rel_work_hybrid} gives an overview of hybrid prediction approaches. Finally, \autoref{subsec:rel_work_summary} closes the section with a brief literature discussion, leading to the contributions of this article in \autoref{subsec:contributions}. 


\subsection{Classification Approaches}\label{subsec:rel_work_classifiers}
Classification approaches for maneuver recognition are described in \cite{weidl2018, wissing2017lane, schlechtriemen2014lane, bahram2016}. In \cite{weidl2018}, a system is introduced, which is capable of detecting lane changes with high accuracies ($>$99\,\%), approximately 1\,s before their occurrence. For this purpose, dynamic Bayesian networks are used. Another approach, which is capable of detecting lane changes approximately 1.5\,s before their occurrence, is presented in \cite{wissing2017lane}. To achieve this, the lane change probability is decomposed into a situation- and a movement-based component, resulting in an $F_1$-score better than 98\,\%. The approach presented in \cite{schlechtriemen2014lane}, in turn, shows that it is possible to detect lane changes up to time horizons of 2\,s when using feature selection for scene understanding, with an Area Under the Curve (\textit{AUC}) better than 0.96. Moreover, \cite{bahram2016} combines interaction-aware heuristic models with an interaction-unaware learned model. The interaction-aware component relies on a multi agent simulation based on game theory, in which each agent simultaneously tries to minimize different cost functions. These cost functions are designed using expert knowledge and consider traffic rules. In a second step, the output of the interaction model is used to condition an interaction-unaware classifier based on Bayesian networks. The approach is able to detect lane changes on average 1.8\,s in advance, with an \textit{AUC} better than 0.93.

\subsection{Trajectory and Position Prediction Approaches}\label{subsec:rel_work_traj_pred}
Approaches dealing with the prediction of trajectories and positions are presented in \cite{lenz2017, altche, wiest2012probabilistic, wiest2013incorporating, schlechtriemen2014probabilistic}: \cite{lenz2017} uses a fully-connected Deep Neural Network to learn the parameters of a two-dimensional \textit{GMM}. For each situation, an adapted Gaussian Mixture distribution models the probability density in the output dimensions $a_x$ and $v_y$ (cf. \autoref{tab:featureOverview}). This distribution is then sampled to estimate trajectories. The authors evaluate their approach with the widely used NGSIM data set \cite{colyar2007us} and show that a root weighted square error (comparable to \textit{RMSE}) of approximately 0.5\,m in lateral direction at a prediction horizon of 5\,s can be achieved.

Another approach, also evaluated with the NGSIM data set, is presented in \cite{altche}. The authors propose the use of a Long Short Term Memory network for predicting trajectories. In particular, the approach is able to compute single shot predictions with an \textit{RMSE} of approximately 0.42\,m at a prediction horizon of 5\,s. \cite{wiest2012probabilistic} deals with the prediction of spatial probability density functions, especially at road intersections. More precisely, a conditional probability density function, which models the relationship between past and future motions, is inferred from training data. Finally, standard \textit{GMM}s and variational approaches are compared. In \cite{wiest2013incorporating}, this approach is extended by a hierarchical Mixture of Experts that allows to incorporate categorical information. The latter includes, for example, the topology of a road intersection. 

In \cite{schlechtriemen2014probabilistic}, a Gaussian Mixture Regression approach for predicting future longitudinal positions as well as a procedure for estimating the prediction confidence are introduced.


\begin{figure*}[!ht]
  \centering
  \includegraphics[scale=0.5]{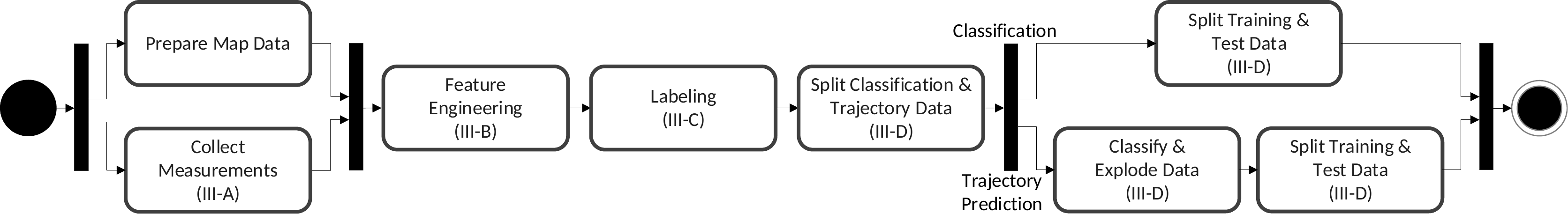}
  \caption{Preprocessing steps used in the proposed workflow (respective sections are referred in the boxes).}
  \label{fig:preprocessing}
\vspace{-4mm}
\end{figure*}

\subsection{Hybrid Approaches}\label{subsec:rel_work_hybrid}
Approaches that combine strategies for both maneuver detection and trajectory or position prediction, similar to the approach presented in this article, are described in \cite{yoon2016, woo2017, wissing2017probabilistic, wissing2018trajectory, deo2018would, deo2018convolutional}. In the following, we denote such approaches as hybrid.

\cite{yoon2016} presents a two-staged approach: In the first step, a Multilayer Perceptron (\textit{MLP}) is used to estimate the future lane of a vehicle. In a second step, a concrete trajectory realization is estimated with an additional \textit{MLP}. As a result, the lane estimation module is able to detect lane changes 2\,s in advance with an \textit{AUC} better than 0.90. The evaluation of the trajectory prediction module shows a median lateral error of approximately 0.23\,m at a prediction horizon of 5\,s.

\cite{woo2017} proposes another hybrid approach that uses the prediction of future trajectories to forecast lane change maneuvers. Moreover, the intention of drivers is modeled using a Support Vector Machine. Subsequently, the resulting action is checked for collisions. This enables the approach to model interrupted lane changes. During the evaluation, an $F_1$-score of 98.1\,\% with a detection time up to 1.74\,s is achieved.

In turn, \cite{wissing2017probabilistic} does not follow such a hybrid approach, but contains an intermediate step before predicting trajectories. Instead of learning maneuver probabilities, the authors present a regression technique for estimating the time span to the next lane change relying on Random Forests. In \cite{wissing2018trajectory}, this approach is extended and combined with findings from \cite{wissing2017lane}. The estimated time up to the next lane changes to the left and to the right are used as input for a cubic polynomial which is intended to predict future trajectories. Finally, the approach is evaluated with the mentioned NGSIM data set, showing a median lateral error of approximately 0.5\,m at a prediction horizon of 3\,s for lane changing scenarios, assuming a perfect maneuver classification.

\cite{deo2018would} proposes the use of a maneuver recognition based on a Hidden Markov Model, distinguishing between ten maneuver classes. Based on this model, a position prediction module, which combines several maneuver specific variational \textit{GMM}s (according to \cite{wiest2012probabilistic}) and an Interacting Multiple Model, which weights different physical models against each other, are implemented. As the approach uses ten maneuver classes and as the errors are only measured in terms of Euclidean distance, the results are difficult to compare with the ones of other approaches. Additionally, the approach is evaluated on a rather small data set. Finally, in \cite{deo2018convolutional} these findings are pursued by the use of a Long Short Term Memory network. The authors demonstrate certain improvements compared to their previous work, while using the NGSIM data set for evaluation purposes.

\cite{schlechtriemen2015will} presents an approach predicting future lateral vehicle positions based on Gaussian Mixture Regression and a Mixture of Experts with a Random Forest as gating network. The approach is evaluated based on a small data set, leading to noisy results, especially in case of lane changes. The evaluation shows that the approach is able to perform maneuver classifications with an \textit{AUC} better than 0.84 and lateral position predictions with a median error of less than 0.2\,m at a prediction horizon of 5\,s.

\subsection{Discussion}\label{subsec:rel_work_summary}

The findings of our literature survey can be summarized as follows: Many works provide meaningful algorithmic contributions. However, in numerous cases we miss structure regarding the problem resolution strategy. Often, it does not become clear how the approaches compare to any baseline (e.\,g. \cite{deo2018would}). Moreover, parameters (e.\,g. \cite{woo2017}) and feature sets (e.\,g. \cite{altche}) are selected manually, and are thus difficult to retrace. In addition, most approaches focus on short or medium prediction horizons (e.\,g. \cite{weidl2018}), or lack a good prediction performance for larger time-horizons (e.\,g. \cite{wissing2018trajectory}). When analyzing the approaches that aim to resolve the long-term prediction problem, it becomes clear that the latter is challenging as the prediction models become significantly more complex as, e.\,g., pointed out by \cite{bahram2016, schlechtriemen2014lane} and \cite{klingelschmitt}. 

Moreover, many approaches (e.\,g. \cite{altche}) aim to predict single trajectories or single shot predictions rather than probabilistic distributions of future vehicle positions. Therefore, the objective to be optimized is mostly the root-mean-square error ($RMSE$). As opposed to these works, we consider the objective of the learning problem as generating an estimator that models a probability distribution of positions reflecting the frequencies of all observed positions, e.\,g., for different drivers in the same situation. Thus, we aim to maximize the likelihood of truly occupied positions given the model. As reasoning behind this design choice, such distributions contain significantly more information than single shot predictions. Thus, they are more useful for applications that need to consider risks, like, for example, maneuver planning approaches as presented in \cite{wiest2012probabilistic, schlechtriemen2016wiggling, sadigh2016planning}.

\subsection{Contributions}\label{subsec:contributions}
The contribution of this article is threefold:

\begin{enumerate}
\item We apply a heuristic-free machine learning workflow to generate a model capable of predicting maneuvers and precise distributions of future vehicle positions for time horizons up to 5\,s (reasonable in terms of comparability). This is achieved with a machine learning workflow that omits any human tuned (hyper-) parameters when constructing the classifiers. Note that this includes all aspects involving feature engineering, labeling, feature selection, and hyperparameter optimization for different classification algorithms. Regarding feature engineering and selection, this means that we construct a data set with a large superset of all features, which are potentially relevant for the problem solution beforehand. Afterwards we select a more or less small feature set that still ensures maximum predictive power through an automated feature selection process.
\item We evaluate the modules for maneuver classification and position prediction, where both parts are not only evaluated separately, as in other works (e.\,g.  \cite{wissing2018trajectory}), but as a combined prediction system as well. This concerns the lateral as well as the longitudinal behavior. In this context, we show that directly feeding the results of the classifier into the regression problem produces results comparable to an Mixture of Experts approach. Additionally, we show that relying on the Markov assumption and not modeling the interactions between the traffic participants explicitly, allows producing superior results compared to existing approaches. As opposed to these works, we integrate the different aspects of behavior prediction, which comprise the prediction of driving maneuvers and positions both in lateral and longitudinal direction. In addition, we introduce new methodologies and conduct a large-scale evaluation.


 \item We demonstrate that the presented methods not only have the potential to outperform state-of-the-art approaches when feeding them with a sufficient number of data. Additionally, we show that our approach is able to provide a meaningful estimate of the prediction uncertainty to the consumer of the information, which is beneficial for collision risk calculation and trajectory planning (e.\,g. \cite{schlechtriemen2016wiggling}).
\end{enumerate}

\section{Data Preparation \& Experimental Setup}\label{sec:pre}
\autoref{subsec:dataset} introduces the considered data set and the experimental setup. \autoref{subsec:features} then gives a detailed overview of the features used to train our models. Afterwards, \autoref{subsec:labeling} introduces the labeling process. Finally, \autoref{subsec:split} deals with the data set split for training, validating and testing the constructed models as well as further preprocessing steps. \autoref{fig:preprocessing} summarizes the overall preprocessing workflow.


\subsection{Data Collection}\label{subsec:dataset}
For modelling and evaluating our modules, we use measurement data from a fleet of testing vehicles \cite{tattersall2012} equipped with common series sensors. The sensor setup includes a front-facing camera detecting lane markings as well as two radars observing the traffic situation in the back. In addition, the vehicles have a front-facing automotive radar to sense the distances and velocities of surrounding vehicles. The data has been collected with different vehicles and drivers at varying times of the day during all seasons. The data collection campaign spanned over more than a year and was mainly restricted to the area around Stuttgart in Germany. Through the wide variance, we are expecting our models to achieve good generalization characteristics.
 
 

Unlike other contributions (e.\,g. \cite{schlechtriemen2015will}), we are not using the actual object-vehicles as prediction target $o$ in this work, but rather the ego- (or measurement-) vehicle itself. However, as our work of course focuses on the prediction of surrounding vehicles, we solely use features that are observable from an external point of view, as postulated in other works (e.\,g. \cite{weidl2018} or \cite{woo2017}). Note that this constraint excludes features like driver status or steering wheel angle. Thus, the models remain applicable to actual object-vehicles, assuming a good sensing of their surrounding. Working with the ego-vehicle data offers several advantages concerning the modeling of situations: First, each situation can be described in a similar way, as situations in which relevant neighboring vehicles to the target-vehicle are hidden for the measurement-vehicle can not occur. In addition, all measurements span longer time periods as the target-vehicle can never disappear from the field of view. This way of data handling is widespread in literature (e.\,g. \cite{wissing2017lane}). In addition, one can expect that future sensor setups will minimize measurement uncertainty for perceived objects and will get closer to the data quality that is nowadays available for the ego-vehicle.

\begin{figure}[!t]
  \centering\includegraphics[height=4.5cm]{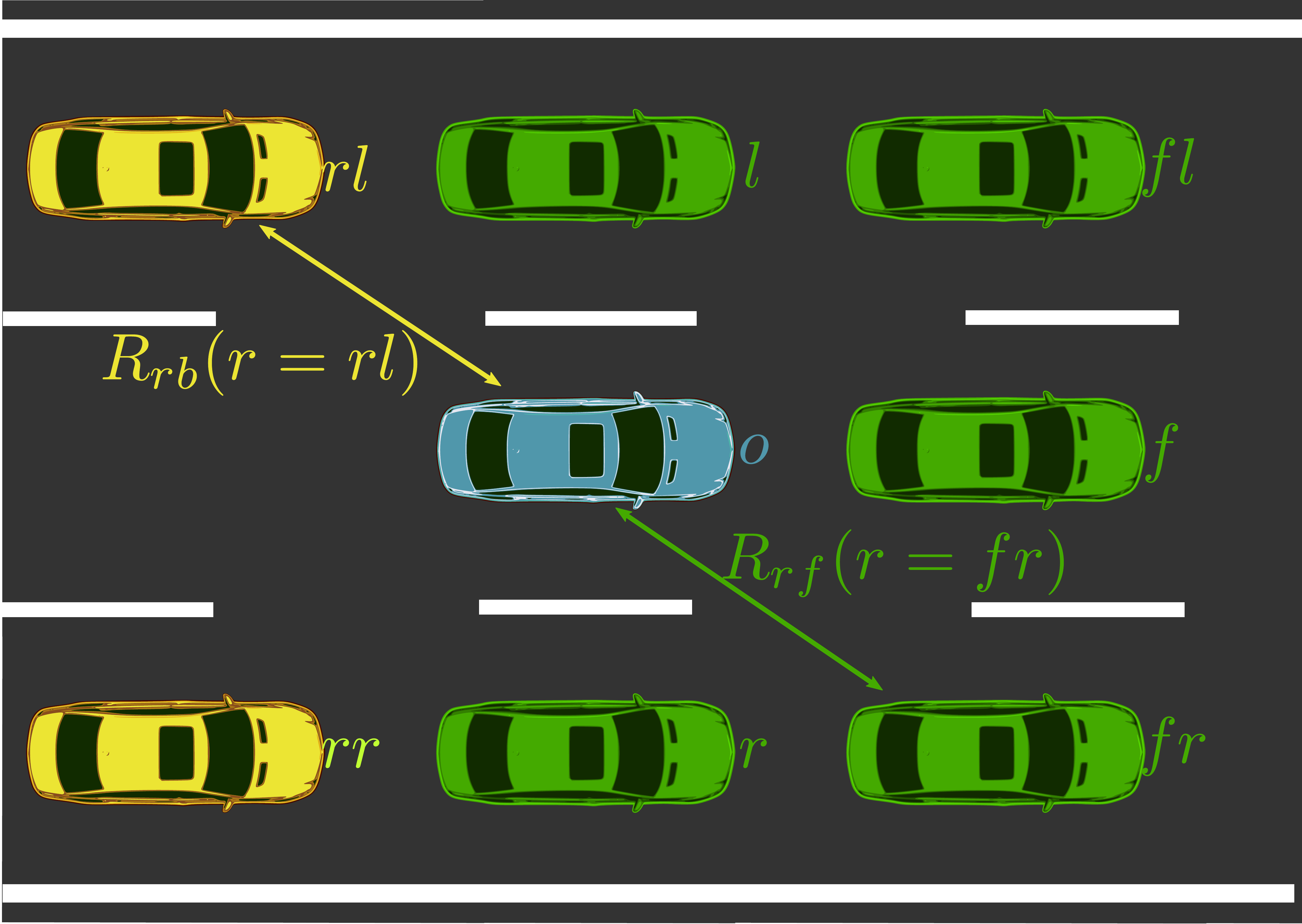}
  \caption{Environment model used for our investigations.}
  \label{fig:env_model}
\end{figure}

Basically, our investigations rely on a similar environment model than the one presented in \cite{schlechtriemen2014lane}, modeling the surrounding with a fixed grid of eight relation partners. But opposed to \cite{schlechtriemen2014lane}, we use the ego-vehicle as prediction target. For this purpose, we slightly adapt the environment model: As the sensors facing the rear traffic in the testing vehicles are less capable than the ones facing the front, our environment model (cf. \autoref{fig:env_model}) distinguishes between relation partners behind (index $rb$) and in front of (index $rf$) the prediction target $o$. Thus, the relation vectors of the rear objects $R_{rb}$ are shortened compared to the ones of the front objects $R_{rf}$. The relation vectors describe the relation between the respective object and the prediction target. Object-vehicles on the same lane as $o$ and driving behind $o$ are left out, as the current sensor setup is not able to sense them. Consequently, a traffic situation can be described by the feature vector $F_{sit}$, which contains the relations of $o$ and its seven relation partners, its own status $F_o$, and the infrastructure description $F_{infra}$ (cf. \autoref{eq:fsit}):

 


\begin{equation}
\begin{aligned}
F_{sit} = & [R_{rf}(r=fl), R_{rf}(r=f), R_{rf}(r=fr),\\
	& R_{rf}(r=l), R_{rf}(r=r), \\
	& R_{rb}(r=rl), R_{rb}(r=rr), \\
	& F_o, F_{infra}]^{T}
\end{aligned}
 \label{eq:fsit}
\end{equation}

A detailed listing of the particular elements of the relation vectors $R_{rf}$ and $R_{rb}$ as well as $F_o$ and $F_{infra}$ can be found in \autoref{tab:featureOverview}.


\subsection{Feature Engineering}\label{subsec:features}

To test and develop our system and to fill the described environment model, we use fused data originating from three different sources:

\begin{enumerate}
\item The basis for our investigations are measurement data produced by the testing fleet (cf. \autoref{subsec:dataset}).
\item As we identified additional features being of interest as inputs beforehand, we fuse the data with information from a navigation map (e.\,g. bridges, tunnels, and distances to highway approaches).
\item Besides, we calculate some higher order features out of the measurements, as e.\,g. a conversion to a curvilinear coordinate-system along the road \cite{thorvaldsson2015}.
\end{enumerate}

\subsection{Labeling}\label{subsec:labeling}

Like previous works \cite{schlechtriemen2015will}, we divide all samples into the three maneuver classes $LCL$ (lane change left), $FLW$ (lane following), and $LCR$ (lane change right) and apply a labeling process that works as follows: First, for each measurement, the times up to the next lane change to the left neighboring lane ($TTLCL$) and to the right one ($TTLCR$) respectively are calculated. This is accomplished by a forecast in time with the distances to the lane markings. As the moment of the lane change, we define the point in time when the vehicle center has just crossed the lane marking. Subsequently, we determine the maneuver labels of each sample based on a defined prediction horizon $T_h$ according to \autoref{eq:labeling}:

\begin{equation}
   L =
   \begin{cases}
      LCL,& \text{if } (TTLCL \leq T_h)\ \land\ \\ 
     & \; \; \; \;(TTLCL < TTLCR) \\
      LCR,& \text{if } (TTLCR \leq T_h)\ \land\ \\
     & \; \; \; \; (TTLCR < TTLCL) \\
      FLW,& \text{otherwise}\\
   \end{cases}
   \label{eq:labeling}
\end{equation}

We decided to use a horizon of 5\,s, as the duration of lane change maneuvers usally ranges from 3\,s to 5\,s (see \cite{woo2017}). Consequently, it is reasonable to label samples only to an upper boundary of 5\,s as potential lane change samples. Additionally, this value is widely used in literature as longest prediction time (e.\,g. \cite{bahram2016, yoon2016} or \cite{woo2017}) and, therefore, it allows for comparability. However, note that this style of labeling might result in decreased performance values, as detections being slightly more than 5\,s ahead of a lane change count as false positives in the evaluation.

\subsection{Data Set Split}\label{subsec:split}
As shown in \autoref{fig:preprocessing}, we split our data into several parts after executing the mentioned preprocessing steps. The first split divides our data into one part for the maneuver classification $D^{Ma}$ and another one for the position prediction $D^{Po}$. This allows us to produce models based on independent data sets. An overview of the splits as well as the respective data set sizes and identifiers is given in \autoref{tab:data}. 

\begin{table*}[!h]
	 \renewcommand{\arraystretch}{1.25}
	\caption{Data Set Identifiers and Sizes}
	\label{tab:data}
	\centering
\begin{tabular}{|p{0.65cm}|p{0.65cm}|p{0.65cm}|p{0.65cm}|p{0.65cm}|p{0.65cm}||p{1.3cm}|p{1.3cm}|p{1.3cm}|p{1.3cm}|}
	\hline
	\multicolumn{6}{|c||}{Maneuver Data:} & \multicolumn{4}{c|}{Position Data:} \\
	\multicolumn{6}{|c||}{$D^{Ma}$} & \multicolumn{4}{c|}{$D^{Po}$} \\
	\hline
	\multicolumn{5}{|c|}{Training \& Validation:} & \centering{Test:} & \multicolumn{3}{c|}{Training:} & \multicolumn{1}{c|}{Test:} \\
	\multicolumn{5}{|c|}{$D^{Ma}_{TV}$} & \centering{$D^{Ma}_{Te}$} & \multicolumn{3}{c|}{$D^{Po}_{T}$} & \multicolumn{1}{c|}{$D^{Po}_{Te}$} \\
	\multicolumn{5}{|c|}{} & & \multicolumn{3}{c|}{130\,623 Trajectories} & \multicolumn{1}{c|}{20\,000}\\
	\multicolumn{5}{|c|}{} & & \multicolumn{3}{c|}{(7\,s; variable sampling time)} & \multicolumn{1}{c|}{Trajectories} \\
	\cline{1-9}
	\centering{$D^{Ma}_{1}$} & \centering{$D^{Ma}_{2}$} & \centering{$D^{Ma}_{3}$} & \centering{$D^{Ma}_{4}$} & \centering{$D^{Ma}_{5}$} & \centering{$D^{Ma}_{6}$} & \centering{$D^{Po}_{T,LCL}$} &  \centering{$D^{Po}_{T,FLW}$} & $D^{Po}_{T,LCR}$ & (5\,s; 10\,Hz) \\
	\multicolumn{6}{|c||}{Samples per Maneuver Class:} & \multicolumn{3}{c|}{Selected\footnotemark[3] Trajectories:} &  \\
	\centering{90\,759} & \centering{87\,499} & \centering{89\,048} & \centering{90\,458} & \centering{92\,669} & \centering{87\,308} & \centering{3\,685} & \centering{6\,037} & \centering{5\,071}& \\
	\hline
\end{tabular}
\end{table*}

%

The first part $D^{Ma}$ is then used as follows: To prepare the training, parametrization and evaluation of the developed classifiers as well as to stay methodically straight, we split data set $D^{Ma}$ once more into six folds\footnote{As shown in the following sections, the amount of folds is a trade-off between computability and correctness}. Thereof we use five folds $D^{Ma}_{TV}$ in \autoref{sec:classification} for the design and parametrization. The remaining fold $D_6^{Ma}=D^{Ma}_{Te}$ is only used for the performance examinations presented in \autoref{sec:class_eval}. The split is performed based on entire situations as described in \cite{schlechtriemen2015will}. This means that the measurements of each situation solely occur in one of the folds. Note that this ensures the absence of unrealistic results, which might occur due to similar samples from the same time series in the evaluation and trainings data otherwise. To achieve an even proportion of the three maneuver classes, we balance the number of samples within each fold by a random undersampling strategy. As the prediction problem is extremely unbalanced, as outlined in \cite{altche}, classifiers would focus on the most frequent maneuver class $FLW$ otherwise. In our case approximately 94\,\% of the data points belong to that class.

In addition, we only take situations into account that were collected continuously up to the prediction horizon of 5\,s. This ensures that the folds are also balanced over time, which constitutes a prerequisite for performing fair evaluations. This is necessary, as the prediction task is obviously much more demanding when predicting a lane change 4\,s in advance instead of 1\,s in advance. Due to this strategy, the numbers of samples in the six folds are slightly different, but we consider this as uncritical. Overall, $D^{Ma}$ contains approximately 8 hours of highway driving of which $\frac{2}{3}$ are collected right during lane changes.

The second data set $D^{Po}$, which serves for the training and evaluation of the position prediction, is processed as follows: Initially, we add the lane change probabilities as estimated by the different classifiers to each sample. Furthermore, we only consider measurements that were collected when the vehicle was manually driven. Note that this restriction is essential as all vehicles of our testing fleet are equipped with an Adaptive Cruise Control (\textit{ACC}) system. Thus, driving in a semi-automated mode is over-represented in our data set compared to reality.\footnote{We do not explicitly filter out \textit{ACC} driving in the data set for maneuver classification, as we can assume that \textit{ACC} is always deactivated during lane changes.}

We further split data set $D^{Po}$ into the subsets $D^{Po}_{T}$ for training and $D^{Po}_{Te}$ for evaluating the position predictions (cf. \autoref{sec:traj_pred} and \autoref{sec:traj_eval}). Afterwards, we expand each data point in $D_T^{Po}$ with the desired prediction outputs, i.\,e., the true positions in $x$ and $y$ direction for all times $t \in T_T=$\{-1.0\,s, -0.9\,s, ..., 6.0\,s\}. Note that the samples with negative times and the ones with times $>$5\,s are needed to train the distributions correctly. Strictly limiting the times to a certain range would generate areas in the data space, which are difficult to represent with \textit{GMM}s due to discontinuities similar to the ones in the probability dimension (cf. \autoref{subsec:Integrated}). To overcome these problems, we integrated a mechanism performing a subsampling between \mbox{-1\,s} and \mbox{0\,s} as well as between \mbox{5\,s} and \mbox{6\,s} according to a Gaussian distribution (percentiles: $P_{50}=0.0\,s$; $P_{-3\sigma}=-1.0\,s$; equivalent between 5 and 6\,s). 

Another mechanism performing a time interpolation ensures that the training data points are distributed continuously along the time dimension. Accordingly, we also have access to prediction times in between our sampling times during the training process. Moreover, the data points in the position test data set $D^{Po}_{Te}$ are expanded with $x$ and $y$ positions as well as corresponding times $t \in T_{Te}=$\{0.0\,s, 0.1\,s, ..., 5.0\,s\}.

\pagebreak

Finally, we 'coil' the two data sets $D^{Po}_T$ \& $D^{Po}_{Te}$ such that each of the newly constructed data points contains the features at the start point of the prediction, one corresponding prediction time, and the actual $x$ and $y$ positions at that point in time (in \autoref{fig:preprocessing} this step is called 'Explode Data'). Hence, our data sets are multiplied by a factor of $|T_T|=71$ respectively $|T_{Te}|=51$ and are structured as described in \autoref{subsec:traj_measures}. Note that $D^{Po}_{T}$ is re-splitted along the maneuver labels and undersampled in \autoref{subsec:MOE}, to train maneuver specific position prediction experts.



\addtocounter{footnote}{1}
\footnotetext[\thefootnote]{for details see \autoref{subsec:MOE}}

\section{Maneuver Classifier Training}\label{sec:classification}
\begin{figure}[!ht]
	\centering\includegraphics[scale=0.41]{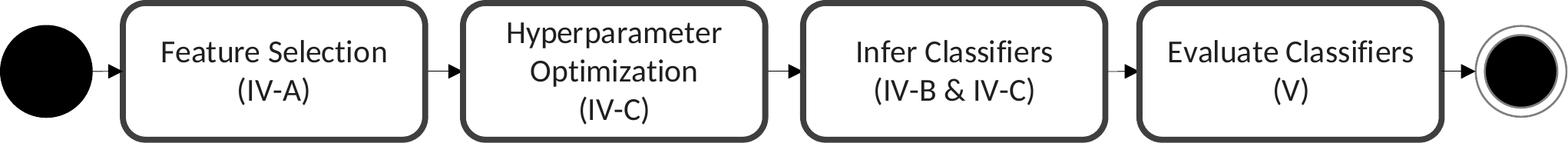}
	\caption{Process of training and evaluating maneuver classifiers.}
	\label{fig:classifier}
\end{figure}

This section gives an overview of the different techniques used for feature selection (cf. \autoref{subsec:feature_selection}), classification algorithms (cf. \autoref{subsec:algorithms}), and techniques to tune the respective hyperparameters (cf. \autoref{subsec:hyperparameter}) for the maneuver classification. The corresponding activities are illustrated by \autoref{fig:classifier}.

\subsection{Feature Selection}\label{subsec:feature_selection}
This section deals with the task of selecting a meaningful subset of features from the available superset. Such selection makes sense for two reasons: First, it can improve the prediction performance of the maneuver classifiers. Second, it can help to reduce calculation efforts, enabling predictions on devices with limited computational power as well. Our main goal here is to improve the overall prediction performance. Note that this slightly contrasts with an overall ranking of the available features, as some of them are highly redundant. Consequently, the most predictive variables shall be selected, while excluding redundant ones. In literature, one can find numerous works dealing with feature selection in machine learning applications. In our implementation, we rely on the findings from \cite{guyon2003}. As we claim to solve the underlying classification problem through a systematic machine learning workflow, we start with simple techniques and move towards more sophisticated and computationally expensive ones. To demonstrate the performance of the used techniques, additionally, we test the classification with the entire superset as a baseline. The superset that contains all features is denoted as $A$ in the following.


\pagebreak

The first investigated technique is a simple correlation-based feature selection technique, which evaluates the correlation of all features and then applies a threshold (set to 0.15) to remove features showing a very low correlation with the maneuver class from the superset. More precisely, we compute Spearman's Correlation (see \cite[p. 133 ff]{fahrmeir2016statistik}) between each feature and the time up to the next lane change ($TTLC$). We selected this quantity instead of the maneuver label, as it enables a smooth fade-out. The resulting feature set is denoted as $B$ in the following. \autoref{tab:selection_techniques} summarizes the examined variants and their abbreviations. Finally, the elements of the resulting feature sets can be found in \autoref{tab:featureOverview}.

\begin{table}[!h]
	\caption{Summary of Examined Feature Selection Techniques}
	\label{tab:selection_techniques}
	\centering
	\begin{tabular}{|c|c|}
		\hline
		Variant & Description\\
		\hline
		$A$ & Superset as Baseline\\
		$B$ & Correlation Threshold\\
		$C$ & \textit{CFS} \\
		$D$ & Wrapper Technique \\
		\hline
	\end{tabular}
\end{table}

The second technique uses the Correlation-based Feature Selection (\textit{CFS}; cf. \cite{hall2000correlation}) and is referred to as $C$ in the following. For this technique, the correlation of entire feature sets instead of single features is calculated. More precisely, for all feature sets $S$, the 'merit' $M_S$, as a measure of the predictive performance, is computed according to \autoref{eq:merit}:
\begin{equation}
{M}_{S} = \frac{n\,\overline{\rho_{cf}}}{\sqrt{n+n(n-1)\overline{\rho_{ff}}}}
\label{eq:merit}
\end{equation}

$n$ describes the number of features and $\overline{\rho_{cf}}$ corresponds to the mean correlation of all features with the class label or, in our case, $TTLC$. Variable $\overline{\rho_{ff}}$, in turn, describes the mean feature-feature inter-correlation of all features within $S$. As can be seen from \autoref{eq:merit}, strongly correlated features in a feature set $S$ minimize $M_S$, whereas a stronger correlation with the class label $\overline{\rho_{cf}}$ maximizes the value of $M_S$. All these computations rely on the assumption that no strong feature inter-correlations are present in the data set, but that instead every relevant feature itself is at least weakly correlated with the class label (see also \cite{hall2000correlation}). To meet the conditions of our data set and to be consistent with variant $B$, we use Spearman's correlation coefficient. As the computation of $M_S$ is not feasible for all possible feature combinations, we use a backward selection strategy that, according to Guyon \cite{guyon2003}, typically provides superior results compared to forward selection. When applying it in our research, we try to minimize the possible shortcomings of the \textit{CFS} by applying cross-validation with the five data folds for training and validation ($D^{Ma}_{TV}$), as described in \autoref{subsec:split}. 

The feature selection techniques described so far are limited in two aspects: Firstly, a proper incorporation of the properties of the used classification algorithm is missing. Secondly, features only being meaningful in combination with others are not considered in feature sets $B$ and $C$. Therefore, when generating feature set $D$, we apply a wrapper feature selection technique as described in \cite{kohavi1997}. As the training of Random Forests already includes an implicit feature selection, we solely focus on wrapper techniques including the other classifiers presented in \autoref{subsec:algorithms}.
The main idea of wrapper techniques is to incorporate the classifier itself as black box into the feature selection process. Within this process the prediction performance on a validation data set is used to determine the best feature set for the respective classifier.  
We build our investigations on a hyperparameter set that was optimized as described in \autoref{subsec:hyperparameter}, whith the feature set of variant $C$  being used for optimization. According to the process for deriving $C$, we perform the search for the most descriptive feature set  with backward elimination. As for each of the approximately 5\,000 possible subsets, a classifier needs to be trained and evaluated, the wrapper technique becomes computationally expensive. To accelerate the computation, we are not performing the validation using cross-validation. Instead, we use one of the data folds  constructed in \autoref{subsec:split} for training ($D^{Ma}_1$) and one for validation ($D^{Ma}_2$).

\subsection{Examined Classification Algorithms}\label{subsec:algorithms}
For the task of maneuver classification, we consider three different algorithms for evaluation purposes, which have been successfully applied in reference works:

\begin{enumerate}
\item The first algorithm is based on a Gaussian Na\"{i}ve Bayes (\textit{GNB}) approach using \textit{GMM}s instead of only using one Gaussian kernel per class and was presented in \cite{schlechtriemen2014lane}.
\item The second algorithm is based on a Random Forest (\textit{RF}) and was presented in \cite{schlechtriemen2015will}. 
\item The third algorithm is based on a Multilayer Perceptron (\textit{MLP}) approach and was presented similiarly in \cite{yoon2016}. As opposed to \textit{GNB} and \textit{RF}, this approach uses scaled features, as suggested by \cite[p. 398 ff]{hastie2001}. In contrast to \cite{yoon2016}, we use a modified labeling and a partly automated strategy to identify an optimal model structure, where we restrict the model to one hidden layer in order to keep the parameter optimization solvable in finite time.
\end{enumerate}



\subsection{Hyperparameter Optimization}\label{subsec:hyperparameter}

To achieve the best possible performance and to enable a fair comparison of the examined classifiers, we optimize their respective hyperparameters. For the \textit{GNB}, this means to find the optimal number of Gaussian kernels $K$ used for each feature and class. A Variational Bayesian Gaussian Mixture Model (\textit{VBGMM}; see \cite{corduneanu2001variational}) is used in this context. This technique was already successfully applied in \cite{wiest2012probabilistic}. The principle behind \textit{VBGMM}s is to fit a distribution of the possible Gaussian Mixture distributions using a Dirichlet process. Hence, this technique ensures that the optimal value for $k$ is determined automatically. 

Regarding \textit{RF} and \textit{MLP} approaches, the parameter optimization is executed for each feature set using a grid-search. This means, that we vary the parameters and calculate for each parameter set a performance value. For the latter, we calculate the average balanced accuracy (see \autoref{subsec:class_measures}) in a leave one out cross-validation manner. Thereby, we use the data of the five data folds for training and validation ($D^{Ma}_{TV}$). The parameters to be optimized are summarized in \autoref{tab:params}.


\begin{table}[!h]
	\caption{Optimized Hyperparameters per Classifier}
	\label{tab:params}
	\centering
	\tymin 40pt
	\begin{tabulary}{0.9\columnwidth}{|C|C|J|}
		\hline
		Classifier & Parameter & Description \\
		\hline
		\multirow{8}{*}{\textit{MLP}}& \multirow{3}{*}{$\alpha$} & \textit{Step size}: Controls how fast the weights of the network are adapted towards the direction of the gradient \\
		\cline{2-3}
		& & \\[-0.27cm]
		& \multirow{3}{*}{$n_{hidd}$} & \textit{Hidden neurons}: Describes the structure of the network as we are only working with one hidden layer \\
		\cline{2-3}
		& & \\[-0.27cm]
		& \multirow{2}{*}{$n_{iter}$} & \textit{Iterations}: Maximum number of training cycles \\
		\hline
		& & \\[-0.27cm]
		\multirow{6}{*}{\textit{RF}}& \multirow{2}{*}{$n_{tree}$} & \textit{Trees}: Number of parallel trees in the forest \\
		\cline{2-3}
		& & \\[-0.27cm]
		& \multirow{2}{*}{$n_{splt}$} & \textit{Splits}: Maximum number of splits in each tree \\
		\cline{2-3}
		& & \\[-0.27cm]
		& \multirow{2}{*}{$n_{smpl}$} & \textit{Samples}: Minimum number of samples necessary for a split \\ 
		\hline
	\end{tabulary}
\end{table}

So far, we constructed different feature sets (cf. \autoref{subsec:feature_selection}) and optimized the hyperparameters for the different classification algorithms (cf. \autoref{subsec:algorithms} \& \autoref{subsec:hyperparameter}). Subsequently, we now execute a second training step with a larger amount of data for all algorithms, using the optimized feature sets and hyperparameters. The enlargement of the data set is achieved using all five folds that we previously used in the cross-validation $D^{Ma}_{TV}$. Note that through this step we derive the final models for the classifier evaluation (cf. \autoref{sec:class_eval}).


\section{Maneuver Classifier Evaluation}\label{sec:class_eval}

This section presents the experimental results obtained with the trained classification models (cf. \autoref{sec:classification}). \autoref{subsec:class_measures} introduces the used performance measures, whereas \autoref{subsec:class_results} presents and discusses the results measured with the constructed test data set (cf. \autoref{subsec:dataset}).

\subsection{Performance Measures}\label{subsec:class_measures}

To be able to assess the performance of the developed classifiers, several metrics are needed, as we are simultaneously focusing on different objectives. Particularly, we are interested in predicting lane changes not only with high accuracies, but also as early as possible in advance of their execution. 

To reflect that, we use the balanced accuracy ($BACC$), which enables us to perform an even weighting of the classification performance for the three maneuver classes. Basically, we use the definition presented in \cite{brodersen2010balanced}, but in a generalized form for multiclass problems (cf. \autoref{eq:baccmulti}):

\begin{equation}
  BACC = \frac{1}{|M|} \cdot \sum_{m \in M} \frac{TP_m}{P_m}
  \label{eq:baccmulti} 
\end{equation}

$M$ is defined according to \autoref{eq:maneuvers}. Moreover, $TP_m$ corresponds to the number of true positives for class $m$ and $P_m$ to the number of samples truly belonging to class $m$ (positives). Thereby, the classifiers assign each sample to the class with the highest probability value.

Additionally, we use the Receiver Operator Characteristic (\textit{ROC}) and Area Under the \textit{ROC} Curve (\textit{AUC}), which both are widely used metrics in this domain (e.\,g. \cite[p. 180 ff]{murphy2012machine}). As opposed to the $BACC$, the \textit{ROC} curve is originally intended to asses binary classifiers. Accordingly, we transform our three-class problem into three binary classification problems. In contrast to the $BACC$, the \textit{ROC} curves constructed this way enable us to show off the classification performance at different working points (WP). For example, this property allows us to assess the performance for the maneuver classes $LCL$ and $LCR$ with more conservative classifier parametrizations and, thus, less false positives. Additionally, the \textit{AUC} helps to analyze the performance at all possible working points at once.

IBesides, metrics which enable us to analyze the technically possible prediction time horizon are needed. As the point in time being referenced in this context is essential and most sources (e.\,g. \cite{weidl2018}, \cite{yoon2016} and \cite{woo2017}) are not very exact in this respect, we introduce the two metrics $\tau_f$ and $\tau_c$ (cf. \autoref{tab:tau}).


\begin{table}[!h]
	\caption{Definition of the Detection Time Metrics}
	\label{tab:tau}
	\centering
	\tymin 40pt 
\begin{tabulary}{0.9\columnwidth}{|C|J|}
	\hline
	Metric & Definition \\
	\hline
	\multirow{3}{*}{$\tau_{f}$} & Time between the vehicle center crosses the centerline and the first detection of the correct maneuver class (as presented in \cite{wissing2017lane}) \\
	\hline	
	& \\[-0.27cm]
	\multirow{5}{*}{$\tau_{c}$} & Time between the vehicle center crosses the centerline and the moment at which the classifier becomes certain about its decision for a specific maneuver class and does not change it till the end of the situation. Note that this is a by far stricter definition than the one of $\tau_f$\\
	\hline
\end{tabulary}
\end{table}

As opposed to the $BACC$ evaluation, for which an unambiguous class assignment becomes necessary, the class assignment is at this point conducted in a way that matches the binary evaluation in the \textit{ROC} curve: For the classes $LCL$ and $LCR$, respectively, we select a binary decision threshold that keeps the false positive rate below 1\%. The resulting working points are presented later on in \autoref{fig:ROC} along with the $ROC$ curves. The detection times calculated this way reflect an evaluation with a limited false positive rate and, hence, at a similar working point for the different classifiers. Note that this ensures a fair evaluation. We decide here for a very low false positive rate as the system should not produce too many lane change detections. Remember that in practice, lane changes occur very rarely compared to lane following.

\subsection{Results \& Discussion}\label{subsec:class_results}

\autoref{tab:classifierOverview} shows the results ($BACC$, $AUC$, $\tau$) for the different classifiers and feature sets measured based on the maneuver test data set $D^{Ma}_{Te}$. Probably, due to the large number of samples, a favorable classifier parametrization and selection seem to have a significantly higher impact on the classification performance than a clever feature selection has. Note that this can be concluded, as the classifiers working with feature sets $B$ and $C$ only perform slightly worse regarding $BACC$ and $AUC$ than the other classifiers. However, applying a feature selection still remains reasonable as it ensures shorter computation times. In addition, the results indicate that the feature selection contributes to an increase of the prediction times in most cases. Note that this does not apply to the \textit{RF} as this classifier performs an implicit feature selection.

\begin{table}[!ht]
	\caption{Summary of Examined Classifiers with Preferred Hyperparameters}
	\label{tab:classifierOverview}
	\centering
	\begin{tabular}{|c|c|c|c|c|c|}
		\hline
		Classi- & Feature & \multicolumn{4}{c|}{Performance on Test Data}\\
		fier & Set & & \multicolumn{3}{c|}{per Class (\textit{AUC}; $\overline{\tau_{f}}$; $\overline{\tau_{c}}$)} \\
		& & BACC & LCL & FLW & LCR \\
		\hline
		\multirow{12}{*}{\textit{GNB}} & & & 0.924 & 0.815 & 0.905   \\
		& $A$ & 0.704 & 2.86$\pm$1.46\,s & - & 2.92$\pm$1.42\,s  \\
		& & & 1.86$\pm$1.40\,s & - & 2.16$\pm$1.25\,s  \\
		\cline{2-6}
		& & & 0.910 & 0.801 & 0.895  \\
		& $B$ & 0.692 & 2.82$\pm$1.38\,s & - & 2.82$\pm$1.32\,s \\
		& & & 1.91$\pm$1.24\,s & - & 2.06$\pm$1.09\,s \\
		\cline{2-6}
		& & & 0.874 & 0.770 & 0.884 \\
		& $C$ & 0.651 & 2.57$\pm$1.31\,s & - & 2.73$\pm$1.31\,s\\
		& & & 1.85$\pm$1.21\,s & - & 1.97$\pm$1.10\,s\\
		\cline{2-6}
		& & & \textbf{0.943} & \textbf{0.864} & \textbf{0.929} \\
		& $\boldsymbol{D}_{GNB}$ & \textbf{0.772} & \textbf{3.26}$\pm$1.28\,s & - & \textbf{3.11}$\pm$1.14\,s\\
		& & & \textbf{2.41}$\pm$1.29\,s & - & \textbf{2.61}$\pm$1.03\,s\\
		\hline
		\multirow{12}{*}{\textit{MLP}}& &  & 0.973 & 0.909 & \textbf{0.961} \\
		& $A$ & 0.822 & 3.67$\pm$1.26\,s& - & 3.35$\pm$1.19\,s\\
		& & & 2.58$\pm$1.51\,s & - & 2.81$\pm$0.99\,s \\
		\cline{2-6}
		& &  & 0.974 & 0.912 & 0.958 \\
		& $B$ & \textbf{0.831} & 3.73$\pm$1.07\,s& - & \textbf{3.60}$\pm$1.15\,s\\
		& & & \textbf{2.72}$\pm$1.40\,s & - & \textbf{2.86}$\pm$1.10\,s \\
		\cline{2-6}
		& &  & 0.966 & 0.892 & 0.953 \\
		 & $C$ & 0.798 & 3.46$\pm$1.07\,s& - & 3.47$\pm$1.11\,s\\
		& & & 2.69$\pm$1.10\,s & - & 2.81$\pm$0.89\,s \\
		\cline{2-6}
		& & & \textbf{0.976} & \textbf{0.915} & 0.960 \\
		& $\boldsymbol{D}_{MLP}$ & \textbf{0.831} & \textbf{3.79}$\pm$1.16\,s& - & 3.33$\pm$1.18\,s\\
		& & & \textbf{2.72}$\pm$1.45\,s & - & 2.68$\pm$0.98\,s \\
		\cline{2-6}
		\hline
		 \multirow{9}{*}{\textit{RF}} & \cellcolor{Petrol} & \cellcolor{Petrol} & \cellcolor{Petrol} \textbf{0.978} & \cellcolor{Petrol} \textbf{0.925} & \cellcolor{Petrol} \textbf{0.968}\\
		  & \cellcolor{Petrol} $\boldsymbol{A}$ & \cellcolor{Petrol} \textbf{0.838}  &  \cellcolor{Petrol} \textbf{3.81}$\pm$1.14\,s & - & 3.60$\pm$1.19\,s\\
		 & \cellcolor{Petrol} & \cellcolor{Petrol} &  \cellcolor{Petrol} \textbf{3.11}$\pm$1.35\,s & - &  \cellcolor{Petrol} \textbf{3.06}$\pm$1.17\,s \\
		 \cline{2-6}
		& &  & 0.976 & 0.918 & 0.959 \\
		& $B$ & 0.834 & 3.73$\pm$1.13\,s& - &  \cellcolor{Petrol} \textbf{3.61}$\pm$1.17\,s\\
		& & & \cellcolor{Petrol} \textbf{3.11}$\pm$1.29\,s & - & \cellcolor{Petrol} \textbf{3.06}$\pm$1.03\,s \\ 
		 \cline{2-6}
		& &  & 0.964 & 0.893 & 0.953 \\
		 & $C$ & 0.799 & 3.45$\pm$1.07\,s& - & 3.49$\pm$1.10\,s\\
		& & & 2.73$\pm$1.14\,s & - & 2.93$\pm$0.92\,s \\ 
		\hline
	\end{tabular}
\end{table}

\begin{figure*}[!ht]
\begin{subfigure}[]{}
\includegraphics[width=0.3\textwidth]{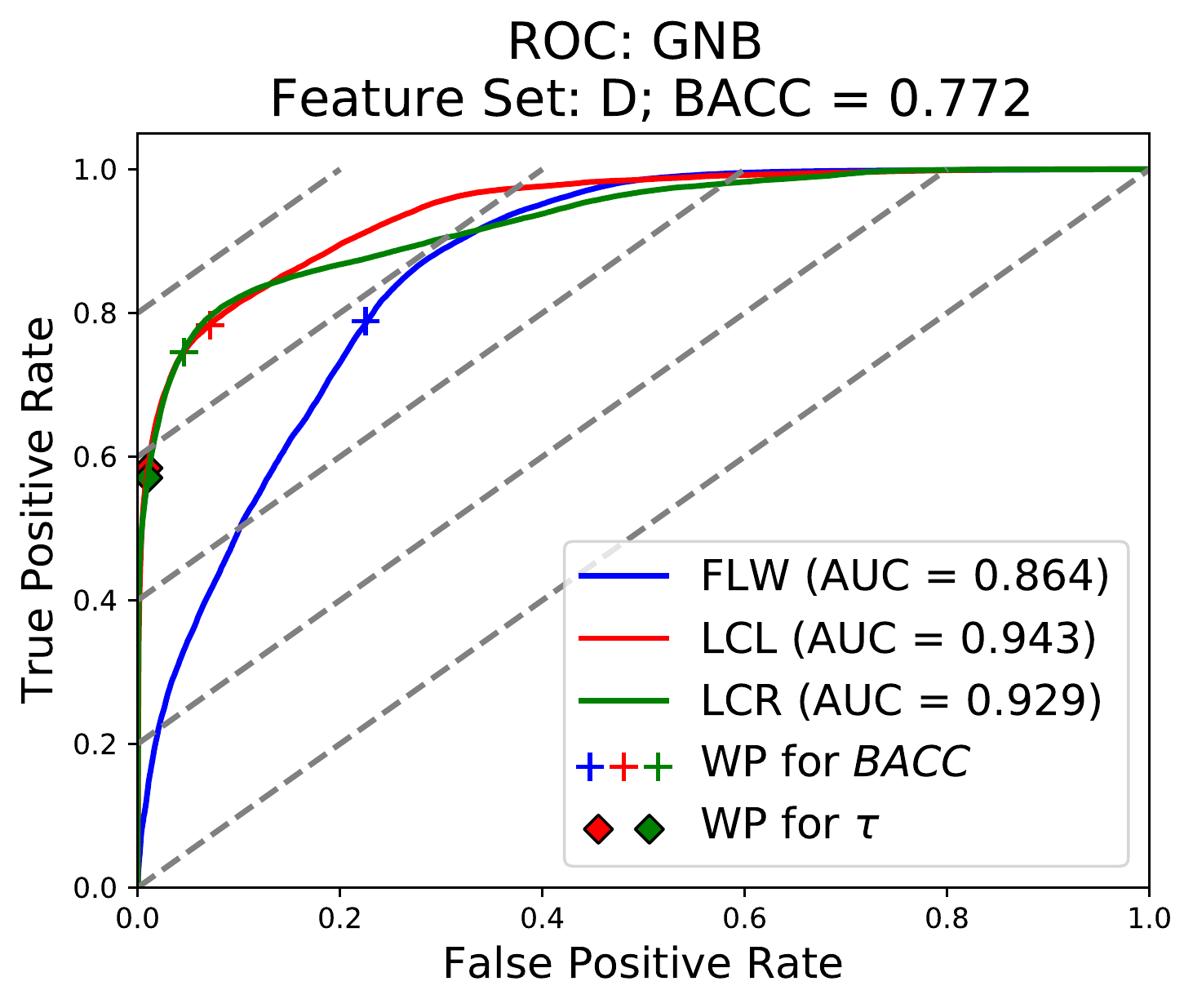}
\end{subfigure}
\begin{subfigure}[]{}
\includegraphics[width=0.3\textwidth]{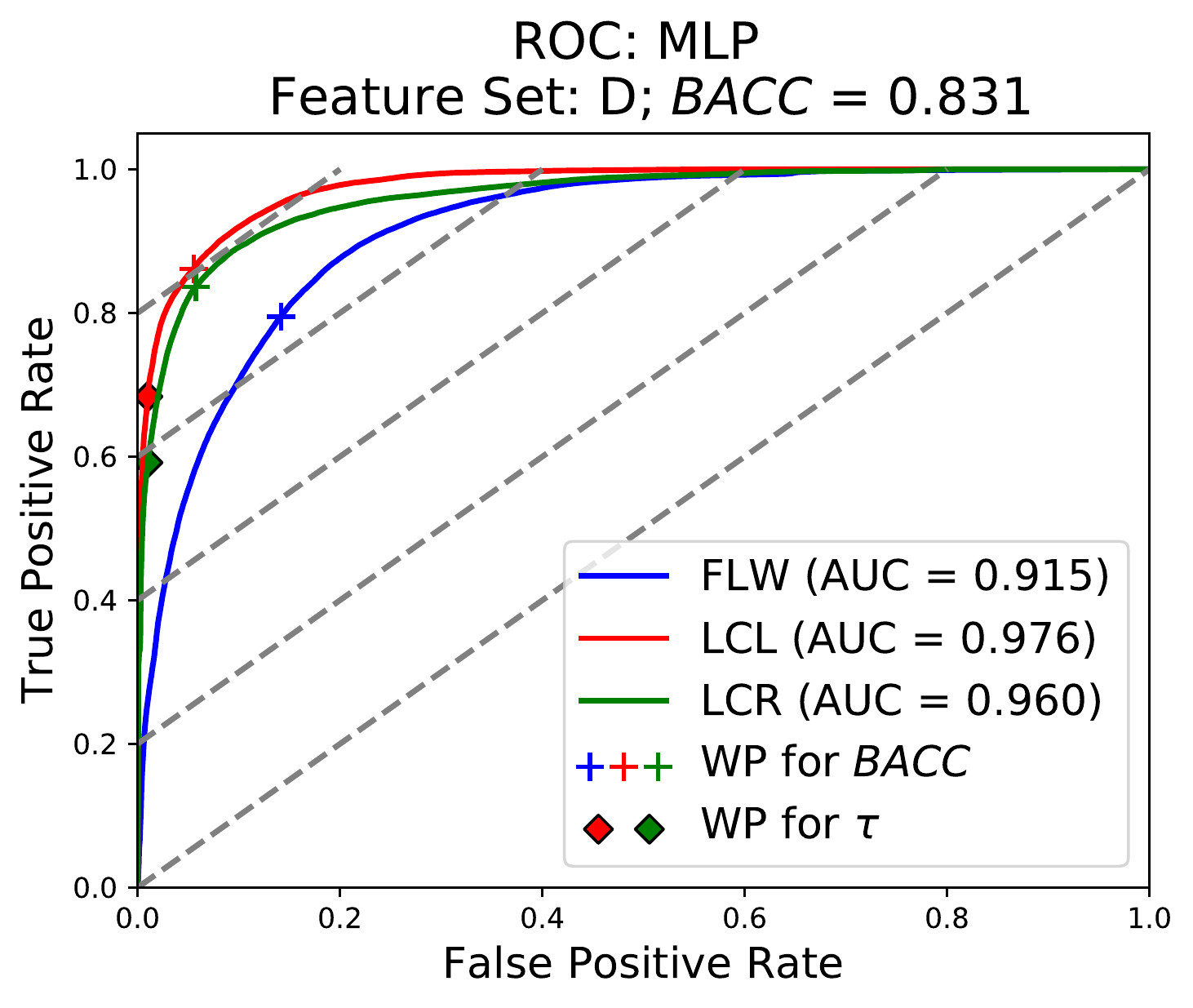}
\end{subfigure}
\begin{subfigure}[]{}
\includegraphics[width=0.3\textwidth]{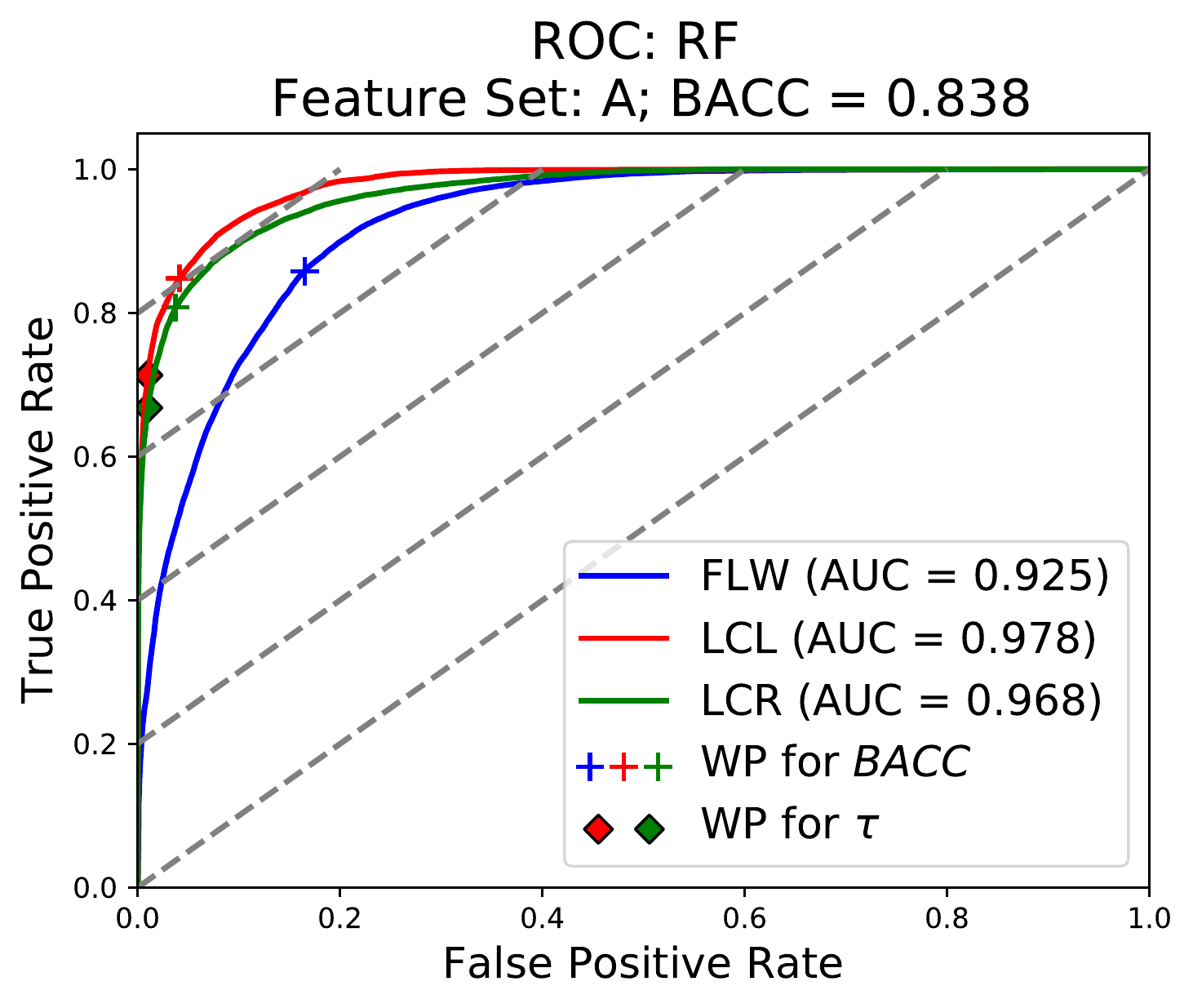}
\caption{\textit{ROC} curves for the developed maneuver classifiers with their respective best parameter sets and hyperparameters.}
\label{fig:ROC}
\end{subfigure}
\end{figure*}

\autoref{fig:ROC} additionally shows the \textit{ROC} curves for the respective best combination of classifier and feature set regarding $BACC$ and $AUC$ for each of the three classifiers. As another result of our investigations, the classification performance for the lane following maneuver ($FLW$), which is neglected by most researchers in literature, is notably worse than for the lane changing maneuvers for all considered algorithms. This can be explained with the fact that nearly each sample, which can not be certainly assigned to one of the lane change maneuvers, is classified as lane following. This is caused, as confusions between a lane change to the right and one to the left are very rare. Thus, a significantly larger number of false positives arises for maneuver class $FLW$. In addition, we could reproduce the findings of \cite{bahram2016}, which showed that lane changes to the left are easier to predict than the ones to the right. One may explain this phenomenon with the observation that lane changes to the right are often motivated by the intention to leave the highway. The latter can be hardly predicted compared to lane changes to the left, which are often performed to overtake slower leading vehicles. Besides, it can be observed that the classification problem remains resolvable even with a significantly decreased number of features, as shown by the \textit{MLP} classifier with feature set $D_{MLP}$, which only includes 24 features. This illustrates that a decreased number of features sometimes leads to an improved performance due to a lower dimension of the input space. This can be explained with the fact that numerous features, which we expected to provide insights into specific lane changing situations, seem to have nearly no effect concerning the general behavior in highway situations. Exemplary features showing this behavior are summarized in \autoref{tab:special_features}.

\begin{table}[!ht]
	\caption{Contextual Features Solely Impacting Special Situations}
	\label{tab:special_features}
	\centering
	\begin{tabular}{|c|c|}
		\hline
		Features & Providing Insights on\\
		\hline
		fog lamps, wiper, ... & weather conditions\\
		\hline
		tunnel, bridge, ... & structural characteristics\\
		\hline
		 lane marking color, ... & road works\\
		\hline
		country, distance to next & \multirow{2}{*}{geographic specialties} \\
		highway exit/approach, ... & \\
		\hline
	\end{tabular}
\end{table}

An explanation of this behavior is that situations, which are affected by these features, occur even rarer than lane changes. However, as automated driving is extremely demanding exactly in these situations, additional investigations are needed in these cases (cf. \autoref{sec:conclusion}).

\begin{figure}[!ht]
\centering
\begin{subfigure}[]{}
\includegraphics[width=0.235\textwidth]{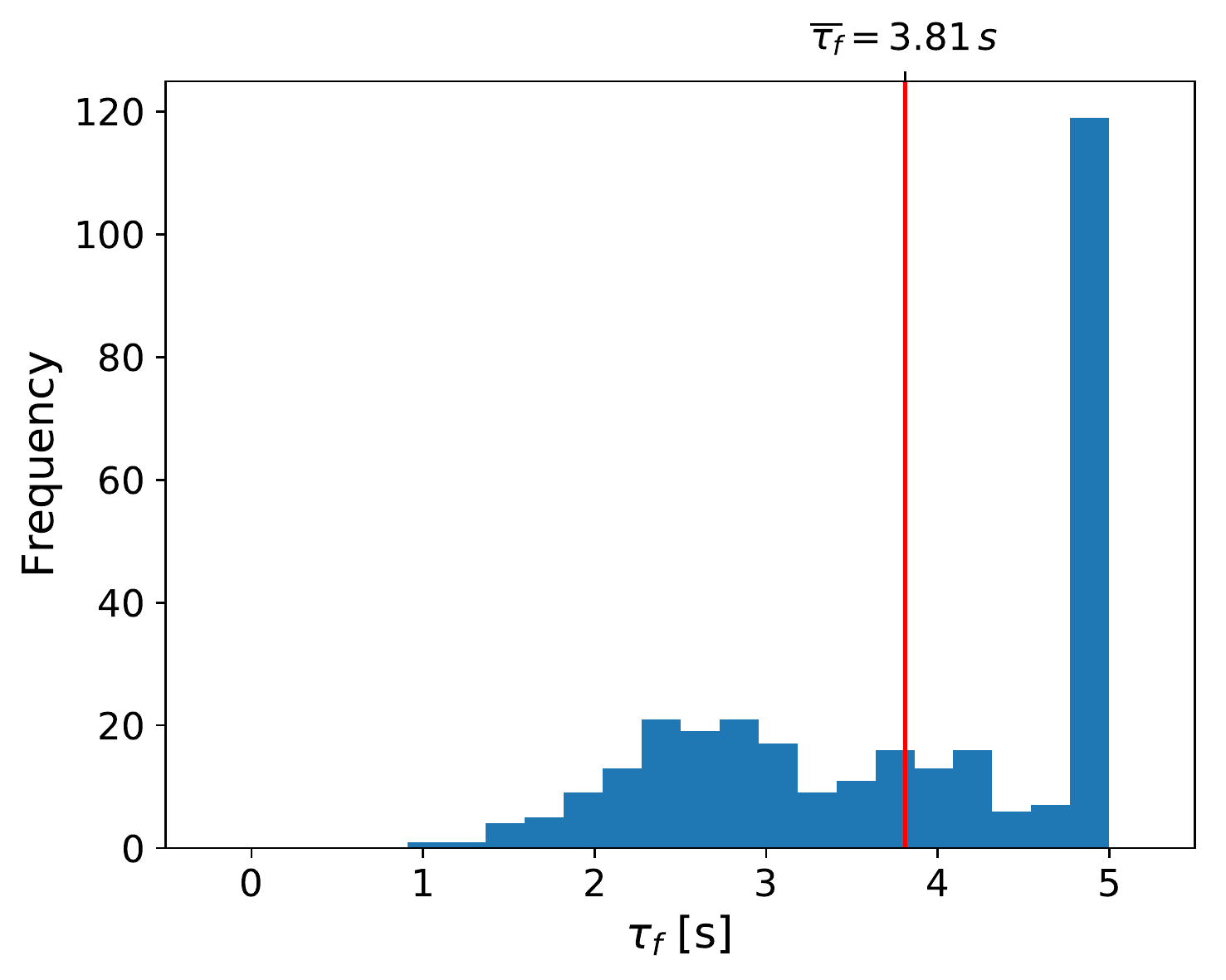}
\end{subfigure}
\hfill
\begin{subfigure}[]{}
\includegraphics[width=0.235\textwidth]{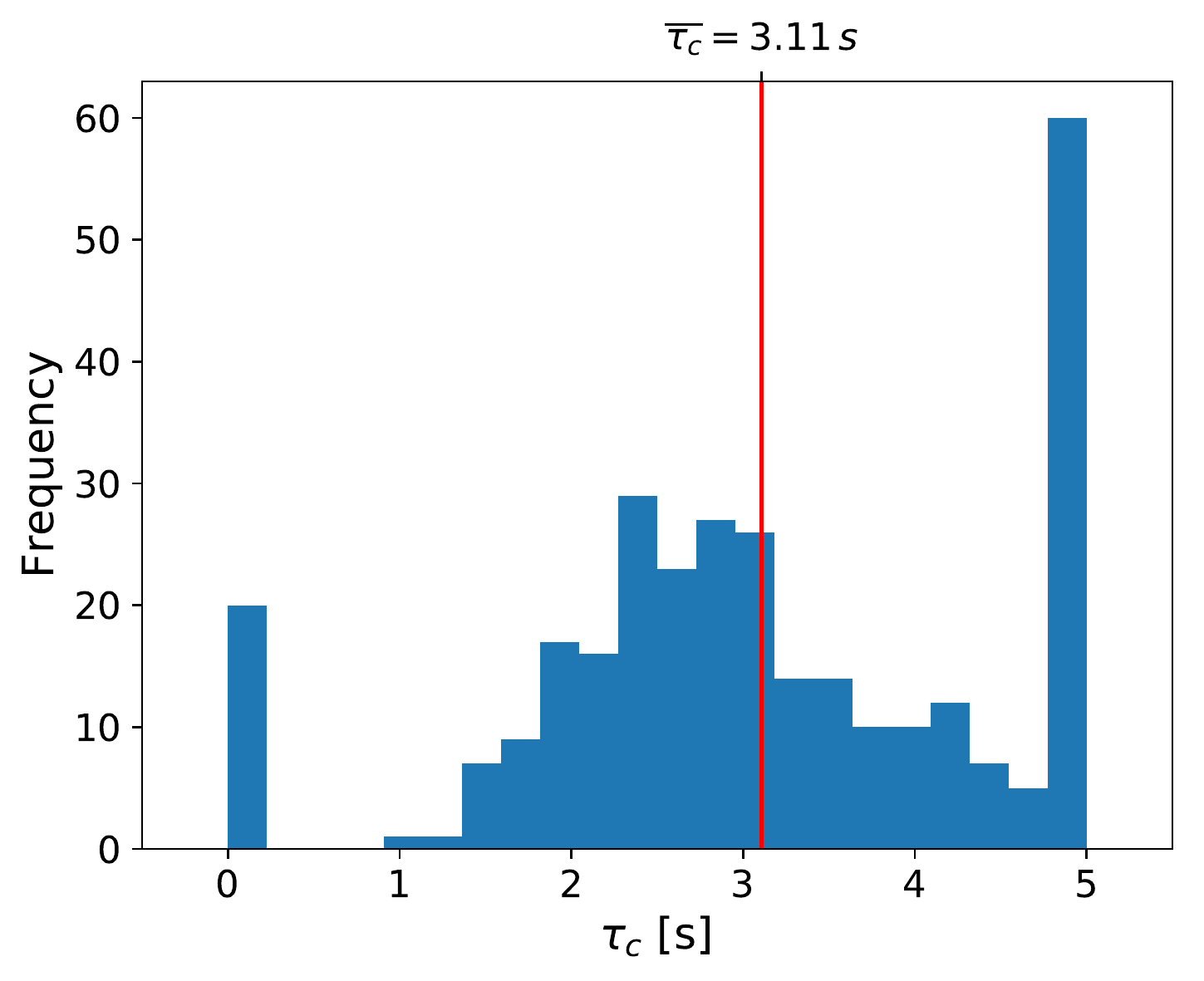}
\end{subfigure}
\caption{Histogram of detection times $\tau_f$ (a) and $\tau_c$ (b) for \textit{RF} for maneuver class $LCL$ with feature set $A$.}\label{fig:hist_tau}
\end{figure}

It is noteworthy that the detection times $\tau_f$ and $\tau_c$ are limited to a maximum of 5\,s due to our evaluation methodology. Therefore, the average values $\overline{\tau_f}$ and $\overline{\tau_c}$ presented in \autoref{tab:classifierOverview} will even be exceeded in practice. To substantiate this assumption, \autoref{fig:hist_tau} shows a histogram of the detection times for the $RF$. The distribution shows numerous situations, that are detected 5 or more seconds in advance.

\pagebreak

Altogether, our investigations show that a systematic machine learning workflow, combined with a large amount of data, is able to outperform current state-of-the-art approaches significantly. This becomes obvious when looking at the \textit{AUC} in comparison to other approaches. \autoref{tab:reference_AUC} shows that our approach outperforms the others, although we are working with a significantly larger prediction horizon, which makes the classification problem more demanding as aforementioned. Finally, note that the mentioned state-of-the-art approaches were designed and evaluated on considerably smaller data sets.

\begin{table}[!ht]
	\caption{\textit{AUC} Values in Comparison to Reference Works}
	\label{tab:reference_AUC}
	\centering
	\begin{tabular}{|c|c|c|c|c|}
		\hline
		Approach & \multicolumn{3}{c|}{\textit{AUC}} & Prediction Horizon\\
		 & LCL & FLW & LCR & \\
		\hline
		\cite{schlechtriemen2015will} & 0.863 & 0.661 & 0.836 & 5.0\,s \\
		\hline
		\cite{schlechtriemen2014lane} & 0.970 & - & 0.991 & 2.0\,s \\
		\hline
		\cite{bahram2016} & 0.947 & - & 0.942 & 2.5\,s \\
		\hline
		\cite{wissing2018trajectory} & 0.934 & - & 0.993 & 2.0\,s \\
		\hline
		\textit{MLP} & 0.976 & 0.915 & 0.960 & 5.0\,s\\
		\hline
		\textit{RF} & 0.978 & 0.925 & 0.968 & 5.0\,s\\
		\hline
	\end{tabular}
\end{table}

Our investigations show that the \textit{GNB} classifier performs significantly worse than the two other approaches (i.\,e. \textit{MLP} and \textit{RF}). Thus, we only use these two classifiers in our further studies. Additionally, we are restricting ourselves to those feature sets and hyperparameter sets showing the best performance (cf. \autoref{tab:selected_clf_params}).

\begin{table}[!ht]
	\caption{Selected Feature Sets and Hyperparameters per Classifier}
	\label{tab:selected_clf_params}
	\centering
	\begin{tabular}{|c|c|c|}
		\hline
		Classifier & Parameter & Value \\
		\hline
		& & \\[-0.27cm]
		 \multirow{4}{*}{\textit{MLP}} & Feature Set & $D_{MLP}$ \\
		 & $\alpha$ & 0.02 \\
		 & $n_{hidd}$ & 27 \\
		 & $n_{iter}$ & 800 \\
		\hline
		& & \\[-0.27cm]
		\multirow{4}{*}{\textit{RF}} & Feature Set & $A$ \\
		& $n_{tree}$ & 128 \\
		& $n_{splt}$ & 16 \\
		& $n_{smpl}$ & 100 \\
		\hline
	\end{tabular}
\end{table}

\section{Position Predictor Training}\label{sec:traj_pred}

\begin{figure}[!ht]
  \centering\includegraphics[scale=0.41]{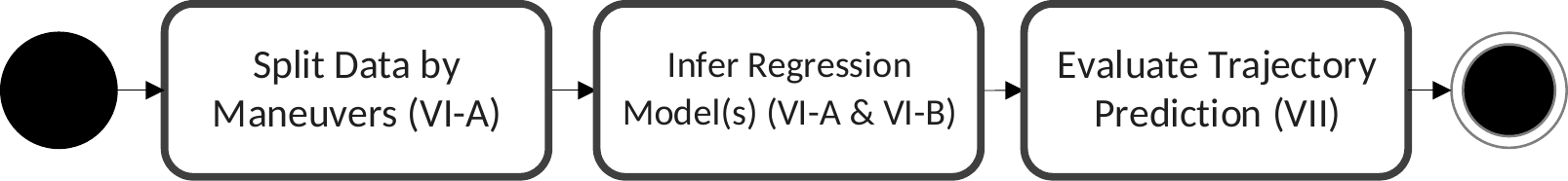}
  \caption{Steps to train and evaluate the position predictors.}
  \label{fig:traj_pred}
\end{figure}

This section deals with the training of the models for position prediction. In particular, we show how to determine the \textit{GMM} parameters $\Theta$. \autoref{subsec:MOE} relies on the Mixture of Experts (\textit{MOE}) approach, which was introduced in \cite{schlechtriemen2015will} for lateral predictions and which uses Gaussian Mixture Regression (cf. \autoref{eq:MOE}). An alternative approach is presented in \autoref{subsec:Integrated}. As opposed to the \textit{MOE} approach, it solves the problem in one processing step (cf. \autoref{eq:IGMM}). The entire procedure, including the evaluation process (cf. \autoref{sec:traj_eval}), is depicted in \autoref{fig:traj_pred}.


\subsection{Mixture of Experts Approach}\label{subsec:MOE}



\begin{figure}[!ht]
  \centering\includegraphics[scale=0.41]{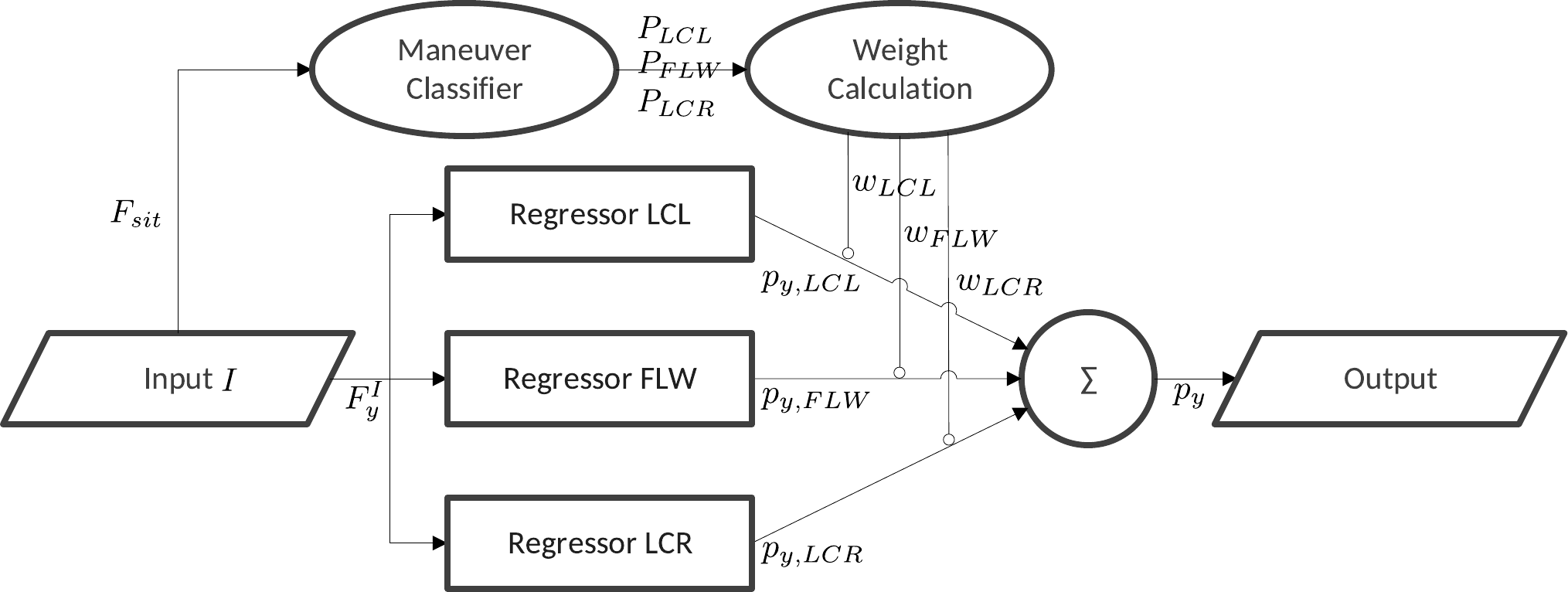}
  \caption{Illustration of the Mixture of Experts (\textit{MOE}) approach.}
  \label{fig:MOE}
\end{figure}

To train the experts for the three maneuver classes, we divide the data set (cf. \autoref{subsec:split}) along the maneuver labels (cf. \autoref{fig:traj_pred}). Subsequently, we perform a random undersampling of the data points for the $FLW$ maneuver class to obtain approximately the same number of samples as for the other two classes. The basic idea behind this step is that the regression problem for the $FLW$ class is less complex than for the two other classes. Thus, it should be solvable with the same amount of data. Amongst others, this data reduction helps to speed up training. As a consequence, the number of $FLW$ samples is approximately decreased by 95\,\% and the data sets $D^{Po}_{T,LCL}$, $D^{Po}_{T,FLW}$, and $D^{Po}_{T,LCR}$ are constructed (cf. \autoref{tab:data}). Afterwards, we train an expert \textit{GMM} with each of these data sets. These experts are later used in the \textit{MOE} approach (cf. \autoref{fig:MOE}). We choose a maximum number of $K=50$ mixture components as well as full covariance matrices\footnote{Preliminary investigations showed that \textit{GMM}s with diagonal covariance matrices are faster to fit, but are by far less accurate.}, and fit the \textit{GMM} in a variational manner again. Besides, we use the following input-feature set $F^I_y$ and the true position $y$ at a defined prediction time $t$ to train the experts in lateral direction (cf. \autoref{eq:Featureset_y}):


\begin{equation}
 F^I_y = \{v_y,\ d_y^{cl}\}
 \label{eq:Featureset_y}
\end{equation}


Regarding the prediction in longitudinal direction, we need to distinguish whether or not a preceding vehicle is present. If no vehicle is in sensor range, both the relative speed and distance for that vehicle are set to default values. As involving the latter in the training of the models would lead to bad fits, the input feature sets $F_{x, Obj}^I$ and $F_{x, \overline{Obj}}^I$ are defined as follows (cf. \autoref{eq:Featureset_x_Obj} \& \autoref{eq:Featureset_x_noObj}):

\begin{equation}
 F_{x, Obj}^I = \{v_x,\ a_x,\ d_v^{rel, f},\ v_v^{rel, f}\}
 \label{eq:Featureset_x_Obj}
\end{equation}

\begin{equation}
 F_{x, \overline{Obj}}^I = \{v_x,\ a_x\}
 \label{eq:Featureset_x_noObj}
\end{equation}

As shown in \cite{schlechtriemen2014probabilistic}, the prediction performance for the longitudinal direction can be significantly increased by learning the deviation from the constant velocity prediction $\hat{x}_{CV}$ instead of the true target position $x$. Consequently, we use the output dimensions $F^O_x$ (cf. \autoref{eq:Featureset_output_x_Obj}): 

\begin{equation}
 F^O_x = \{x-\hat{x}_{CV},\ t\}
 \label{eq:Featureset_output_x_Obj}
\end{equation}
 
\subsection{Integrated Approach}\label{subsec:Integrated}

\begin{figure}[!t]
  \centering\includegraphics[scale=0.41]{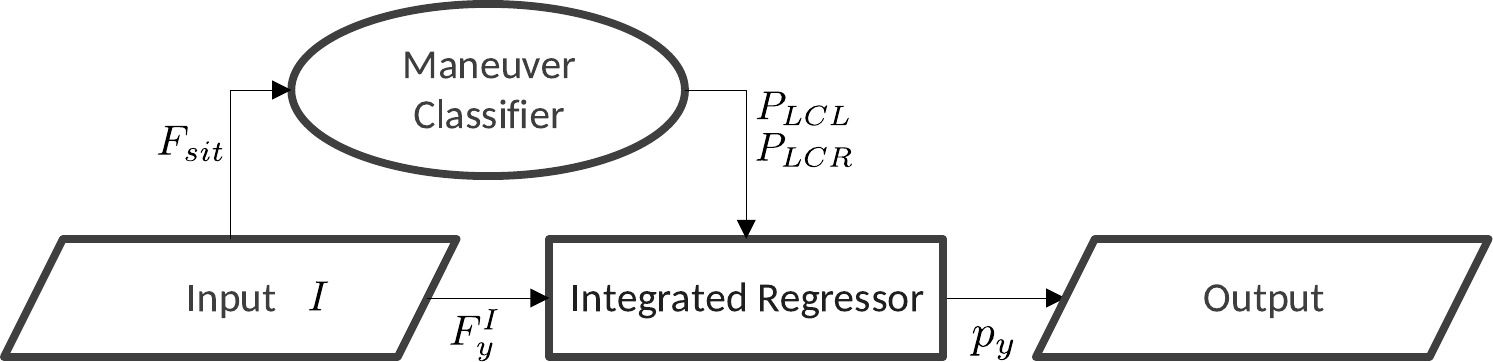}
  \caption{Illustration of the integrated approach.}
  \label{fig:Integrated}
\end{figure}
\begin{figure}[!ht]
\centering
\begin{subfigure}[]{}
\includegraphics[width=0.23\textwidth]{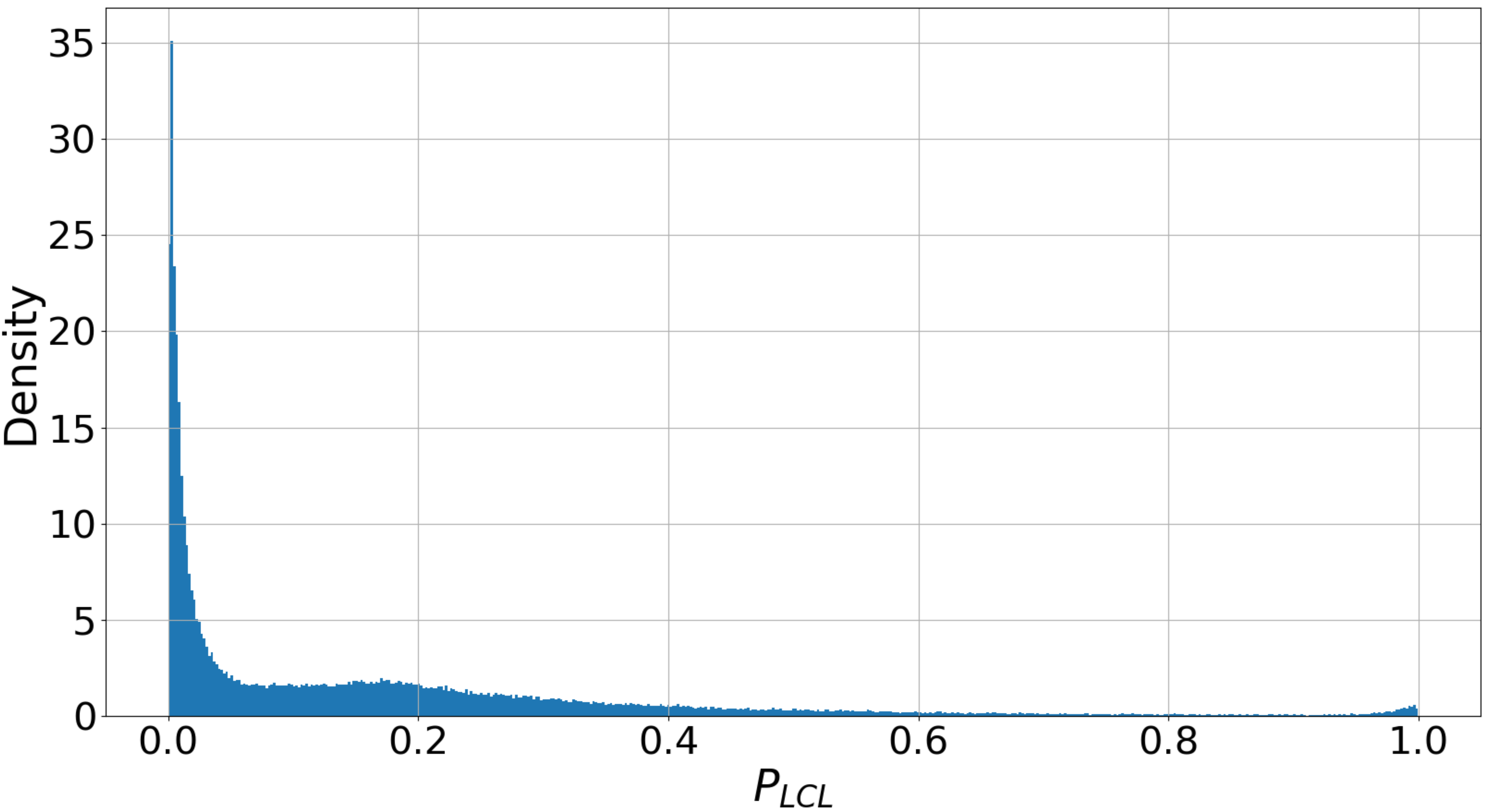}
\end{subfigure}
\hfill
\begin{subfigure}[]{}
\includegraphics[width=0.23\textwidth]{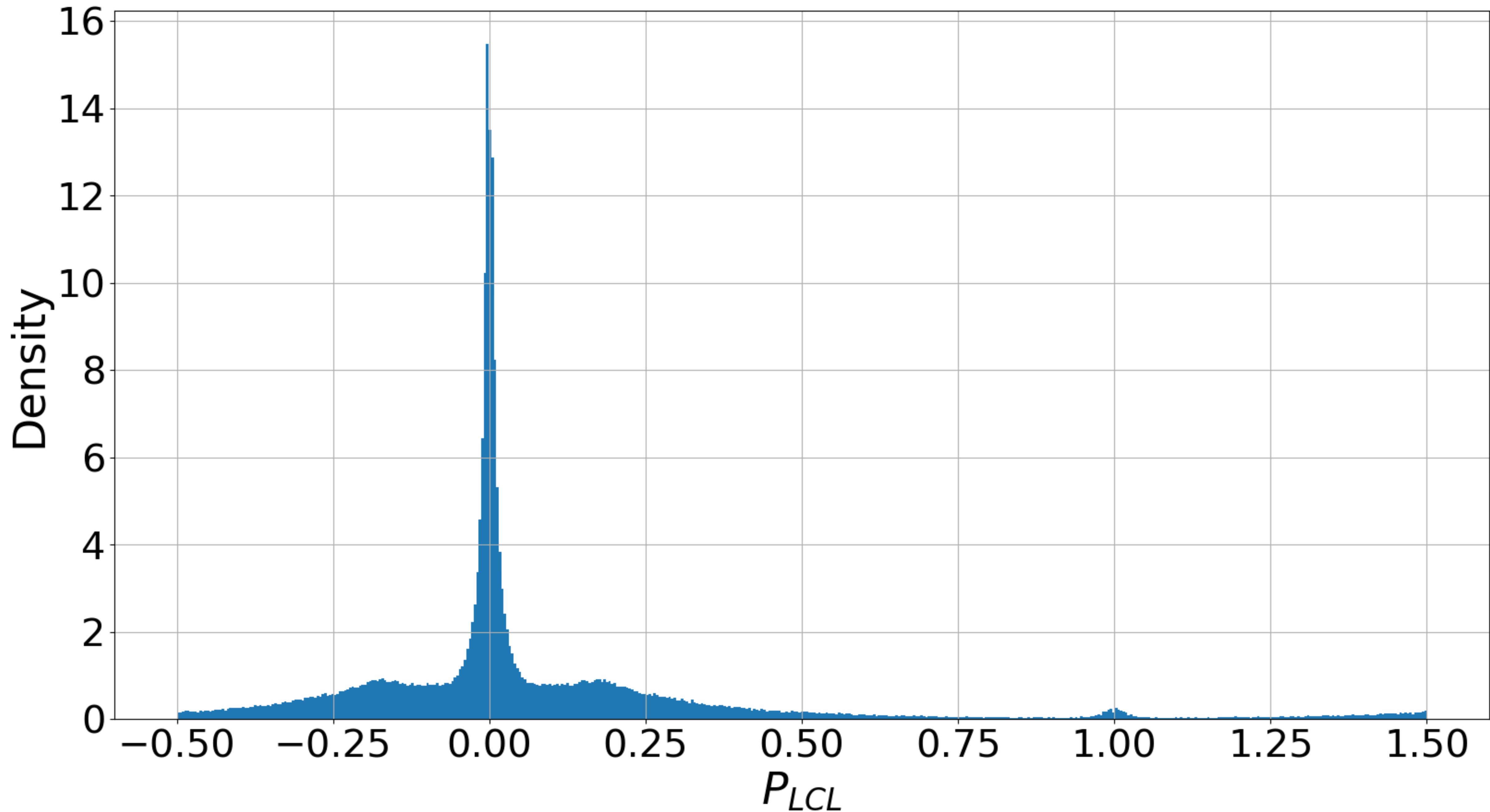}
\end{subfigure}
\caption{Density of $P_{LCL}$ before (a) and after (b) adjustment.}\label{fig:density_plcl}
\end{figure}

As alternative to the \textit{MOE} approach, this section presents an integrated approach, which uses the unsplitted data set $D^{Po}_T$ (cf. \autoref{tab:data}) and expands the feature sets ($F_{x, Obj}^I, F_{x, \overline{Obj}}^I, F_{y}^I$) with the maneuver probabilities $P_{LCL}$ and $P_{LCR}$ (cf. \autoref{fig:Integrated}). $P_{FLW}$ is left out here as this information would be redundant to the one provided by $P_{LCL}$ and $P_{LCR}$, and we want to keep the models' dimension as low as possible. Consequently, the task of considering the maneuver probabilities is directly integrated in the model. The resulting one-block solution is both easier to implement and to use. In this context, we discovered that \textit{GMM}s are not well suited to fit probabilities bounded to values between 0 and 1. Especially, this is the case if most of the probabilities tend against the extreme values (cf. \autoref{fig:density_plcl} (a)). Hence, we expand our data set with a duplicate of each data point containing probability values, which are mirrored at 0 for original probabilities being lower than 0.5 and at 1 for all other original probabilities. This way, we are able to generate the density shown in \autoref{fig:density_plcl} (b), which we identified as easier to fit with \textit{GMM}s. Note that before our adjustment, the density contained an abrupt jump, especially at $P_{LCL}=0$. As such discontinuities are only representable by numerous Gaussian components, which are symmetrical and smooth per definition, many components needed in other areas of the data space would be wasted for this purpose. 

The actual training of the integrated \textit{GMM} is performed similarly to the experts training in a variational fashion, with $K=50$ components and full covariance matrices, but with the entire training data set. Thus, no undersampling procedures are applied and the unbalanced nature of the maneuver classes and their actual frequencies are preserved.

\section{Position Estimation Evaluation}\label{sec:traj_eval}
In order to evaluate the position predictions, first of all, one has to decide which of the considered classifiers fits best as gating network in the Mixture of Experts (\textit{MOE}) and in the integrated approach respectively. Hence, we calculate the average log-likelihoods $\overline{\mathcal{L}}$ on the entire position test data set $D^{Po}_{Te}$ (cf. \autoref{subsec:split}). Note that this data set is not balanced according to the maneuver labels as also suggested in \cite{deo2018convolutional}. In particular, the unbalanced nature of the data allows us to draw general conclusions about the performance, independent of the respective driving maneuver. In this context, the use of the average log-likelihood as quality criterion for comparing different approaches is beneficial, as it rates the quality of the predicted probability density distribution instead of assessing only the ability to predict one single position with maximized accuracy. Moreover, the log-likelihood is exactly the value to be maximized in the process of fitting a \textit{GMM}. However, as  $\overline{\mathcal{L}}$ can not be interpreted as physical quantity, it is solely useful for comparison purposes. As we are also interested in assessing the performance concerning the spatial error and to achieve comparability, we additionally investigate this quantity for the approach working best in the following subsections.

\autoref{tab:likelihoods} shows the per sample log-likelihood of different approaches for the longitudinal ($\overline{\mathcal{L}_x}$) as well as the lateral ($\overline{\mathcal{L}_y}$) direction. In this context, we use the already introduced classifiers \textit{RF} and \textit{MLP} in combination with four different strategies to combine the experts' position estimates, as introduced in \autoref{eq:MOE}, as weighting function $w_m(I)$:

\begin{enumerate}
\item Raw probabilities (Raw): This strategy directly uses the raw probabilities as issued by the classifiers $P_m^{clf}(I)$ as gating probabilities. This means that we concatenate the three \textit{GMM}s and multiply the mixture weights with the probabilities issued by the respective classifier: $w_{m}^{Raw}(I) = P_m^{clf}(I)$.
\item Winner Takes it All (WTA): This strategy uses the outputs of the \textit{GMM} for the maneuver class with the largest probability according to the respective classifier (cf. \autoref{eq:w_wta}).
\end{enumerate}

\begin{equation}
w_{m}^{WTA}(I) = 
   \begin{cases}
      1,& \text{if }P_m^{clf}(I)=\max\limits_{\{q \in M\}} P_{q}^{clf}(I)\\ 
      0,& \text{else}
   \end{cases}
\label{eq:w_wta}
\end{equation}

\begin{enumerate}
\setcounter{enumi}{2}
\item Prior Weighted Raw probabilities (PW-Raw): This strategy considers that the classifiers were trained on a balanced data set. Thus, it multiplies the raw probabilities with the prior probabilities for each maneuver class: $w_{m}^{PWRaw}(I) = norm(P_m^{clf}(I) \cdot \pi_m)$.
\item Integrated \textit{GMM} (I-GMM): This strategy directly uses the integrated approach presented in \autoref{subsec:Integrated} to predict the probability distributions and follows \autoref{eq:IGMM}.
\end{enumerate}

To demonstrate the benefits of our approach, which combines maneuver classification and position prediction, we additionally analyze its performance compared to reference strategies. First, we use the labels as a perfect classifier according to \autoref{eq:labels}:

\begin{equation}
w_{m}^{Labels} = 
   \begin{cases}
      1,& \text{if } m=L\\ 
      0,& \text{else}
   \end{cases}
\label{eq:labels}
\end{equation}

Moreover, we use the pure prior probabilities \mbox{($\pi_{LCL}=\pi_{LCR}=0.03; \pi_{FLW}=0.94$)} as most naive classifier ($w_m^{Priors}=\pi_m$) and a strategy without a classifier, referred to as NOCLF in the following.


\begin{table}[!ht]
	\caption{Per Sample Log-Likelihoods with Different Classifiers and \textit{MOE} strategies}
	\label{tab:likelihoods}
	\centering
	\begin{tabular}{|c|c|c|c|}
		\hline
		&&&\\[-0.8em]
		Classifier & Strategy & $\overline{\mathcal{L}_x}$ & $\overline{\mathcal{L}_y}$ \\
		 &  & (normalized [\%]) & (normalized [\%]) \\
		\hline
		Labels & - & -14.066 (100) & -7.547 (100) \\
		\hline
		Priors & - & -13.273 (106.0) & -7.769 (97.1) \\
		\hline
		NOCLF & - & \cellcolor{Petrol} \textbf{-13.171 (106.8)} & -7.762 (97.2) \\
		\hline
		\multirow{ 4}{*}{\textit{MLP}} & Raw & -13.667 (102.9) & -7.900 (95.5) \\
          	& WTA & -16.279 (86.4) & -8.793 (85.8)  \\
		& PW-Raw & \textbf{-13.329 (105.5)} & \cellcolor{Petrol} \textbf{-7.608 (99.2)} \\
		& I-GMM & -13.354 (105.3) & -7.691 (98.1)\\
		\hline
		\multirow{ 4}{*}{\textit{RF}} & Raw  & -13.568 (103.7) & -7.781 (95.9) \\
		 & WTA & -15.685 (89.7) & -8.369 (90.2) \\
		 & PW-Raw & -13.280 (105.9) & -7.626 (99.0) \\
		 & I-GMM &  \textbf{-13.207 (106.5)} & \textbf{-7.611 (99.2)} \\
		 \hline
	\end{tabular}
\end{table}
%

For the longitudinal direction, \autoref{tab:likelihoods} shows that the reference solution without any previous maneuver classification (\textit{NOCLF}) is able to produce slightly better results than the other combinations. Although it seems to be trivial that lane changes have not to be taken into account when predicting the longitudinal behavior, this is noteworthy, as our expectations beforehand was that lane changes to the left mostly go along with an acceleration, whereas braking actions are extremely rare.


\begin{figure*}[!ht]
  \centering
  \includegraphics[width=0.85\textwidth]{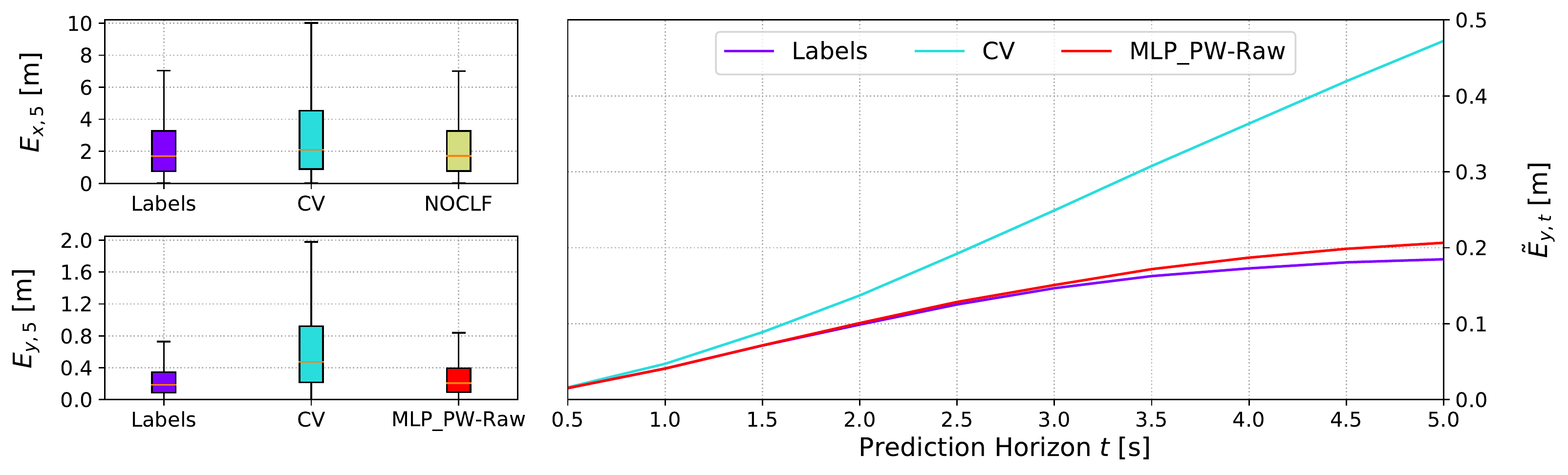}
  \caption{Visualization of the error distribution (left) in longitudinal and lateral direction and the median lateral error as function of the prediction time (right).}
  \label{fig:err_boxplots}
\vspace{-4mm}
\end{figure*}

\pagebreak

By contrast, the benefits of the Mixture of Experts (\textit{MOE}) approach come into effect for the lateral direction. As shown in \autoref{tab:likelihoods}, the combination of prior weighting and \textit{MLP} probabilities performs best. Furthermore, all combinations involving the integrated approach perform only slightly worse or even better (\textit{RF}) than the combinations using prior weighted probabilities. As benefit, these models are easier to use and are more robust against poor or uncalibrated maneuver probabilities without needing an additional calibration step. This can be explained with the fact that these models perform an implicit probability calibration during the training of the \textit{GMM}.



Moreover, we learned that the WTA strategy has no practical relevance, as it does not necessarily produce continous position predictions over consecutive time steps as accomplished by the other strategies per definition. Besides, in case of a misclassification, the WTA strategy solely asks one specific expert model, which might not be applicable in that area of the data space, what clearly decreases the overall performance.

In the following, we investigate the spatial errors of the best combinations (lateral: \textit{MLP} classifier with PW-Raw strategy; longitudinal: \textit{NOCLF}), as previously introduced. For this purpose, we present the applied performance measures in \autoref{subsec:traj_measures} and then show the obtained results in \autoref{subsec:traj_results}.

\subsection{Performance Measures}\label{subsec:traj_measures}

To measure the spatial performance of our predictions, we rely on the unbalanced position evaluation data set $D^{Po}_{Te}$. The latter contains the needed inputs for the maneuver classifiers and position predictors ($I$) as well as the true trajectories $TR$ according to \autoref{eq:eval_data}.

\begin{equation}
  D^{Po}_{Te} = 
    \begin{bmatrix}
      I & TR
    \end{bmatrix}
     \label{eq:eval_data}
\end{equation}

$TR$ contains $N=20\,000$ 5\,s-trajectories sampled with 10\,Hz (hence 1\,000\,000 samples) according to \autoref{eq:trajectory_data}:

\begin{equation}
  TR = 
  \begin{bmatrix}
      tr^0 &  tr^1 & \hdots & tr^{N}
    \end{bmatrix}
  \label{eq:trajectory_data}
\end{equation}

Each trajectory $tr^i$ consists of 51 corresponding $x$ and $y$ positions, according to \autoref{eq:trajectory}:

\begin{equation}
  tr^i = 
  \begin{bmatrix}
      x^i_{0.0} &  y^i_{0.0} \\
      x^i_{0.1} &  y^i_{0.1} \\
      \vdots  & \vdots \\
      x^i_{5.0} &  y^i_{5.0} \\
    \end{bmatrix}
  \label{eq:trajectory}
\end{equation}

The predicted trajectories $\hat{TR}$ are then calculated with the described classifiers and position predictors in the same format as $TR$. However, as the Gaussian Mixture Regression originally produces probability densities instead of point estimates, these have to be calculated first. This is accomplished by calculating the center of gravity of the density as described in \cite{schlechtriemen2015will}. Accordingly, the prediction error $e^i_t$ of a specific prediction time $t$ for one of the $i$ trajectories is calculated separately for the two dimensions $x$ and $y$ as follows (\autoref{eq:traj_err}):
 
\begin{equation}
  e^i_t = 
  \begin{bmatrix}
      e^i_{x, t} &  e^i_{y, t} \\
    \end{bmatrix}
    =   
    \begin{bmatrix}
      |x^i_{t} - \hat{x}^i_{t}| &  |y^i_{t} - \hat{y}^i_{t}| \\
    \end{bmatrix}
  \label{eq:traj_err}
\end{equation}

Variables $\hat{x}$ and $\hat{y}$ describe the estimated positions, whereas $x$ and $y$ correspond to the actual ones. The individual errors $e^i_t$ of all trajectories $i$ are concatenated to $E_t$ (cf. \autoref{eq:traj_err_overall}):

\begin{equation}
  E_t = 
  \begin{bmatrix}
      E_{x, t} &  E_{y, t} \\
    \end{bmatrix}
     = 
  \begin{bmatrix}
      e^i_{x, t} &  e^i_{y, t} \\
    \end{bmatrix}_{\forall i}
  \label{eq:traj_err_overall}
\end{equation}

At this point, we want to re-emphasize, that although this way of evaluating the performance produces easy to interpret results, it disregards that our original outputs (i.\,e. spatial probability densities) contain much more information than a single point estimation.

%
%
%
%


\subsection{Results \& Discussion}\label{subsec:traj_results}

 
 \begin{figure*}[!ht]
  \centering
  \includegraphics[width=0.63\textwidth]{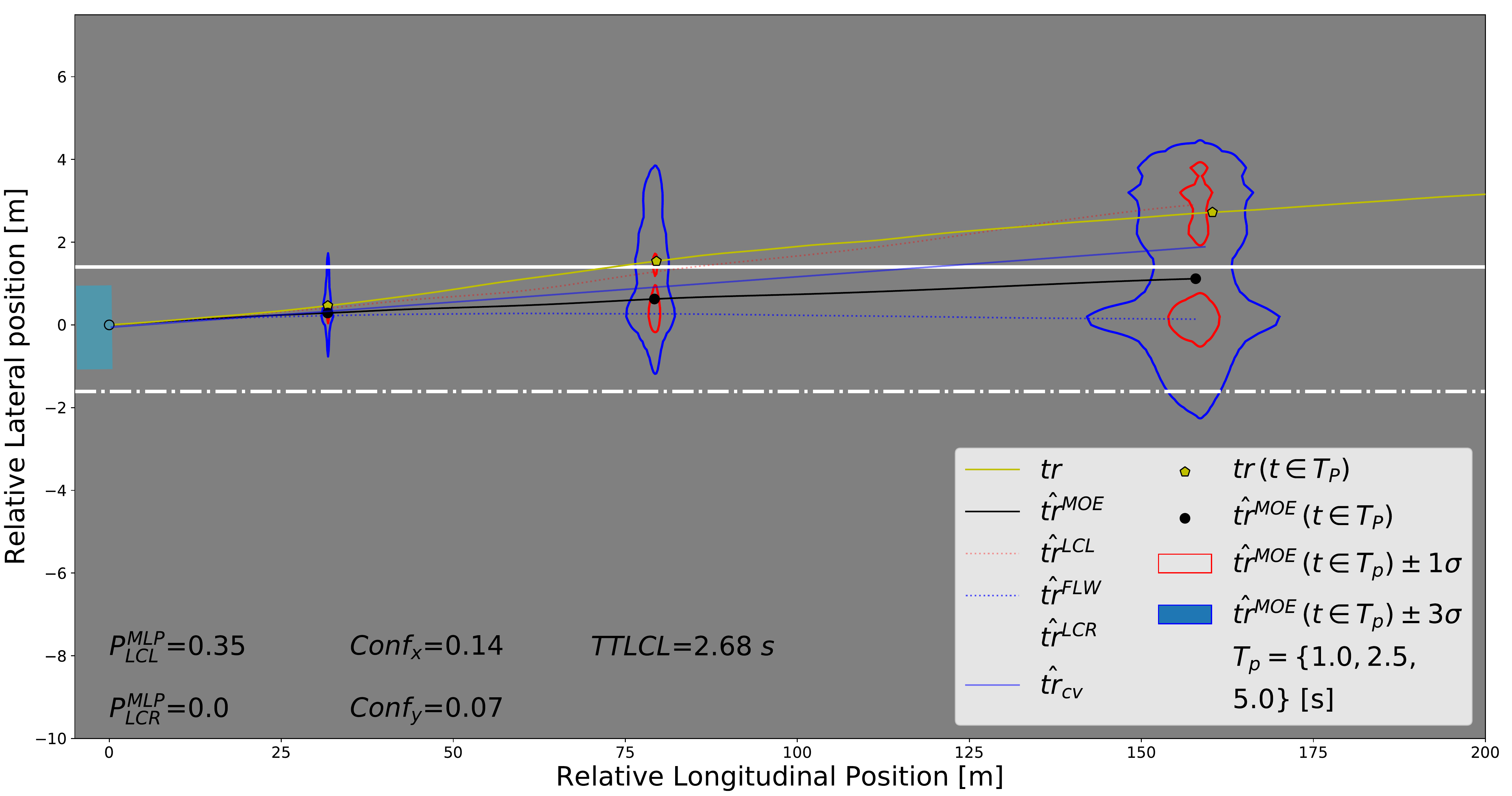}
 \caption{Predicted probability distribution of future vehicle positions for an illustrative situation.}
 \label{fig:predicted_distribution}
\end{figure*}
 
\autoref{fig:err_boxplots} shows the performance of the selected combinations of classifiers and mixing strategies (highlighted in \autoref{tab:likelihoods}) at a prediction horizon of 5\,s for the longitudinal ($E_{x, 5}$) and the lateral ($E_{y, 5}$) direction on the left side. In comparison, a constant velocity (\textit{CV}) prediction and a Mixture of Experts (\textit{MOE}) with labels\footnote{Using the \textit{MOE} with the labels as input corresponds to the assumption of a perfect classifier.} are shown. The right-hand side of  \autoref{fig:err_boxplots} shows the development of the median lateral error $\tilde{E}_{y,t}$ as function of the prediction time $t$.

As the plots indicate, our position prediction system is able to produce results comparable to the ones with a perfect maneuver classification, in both lateral and longitudinal direction. Additionally, the plots show that we are able to clearly outperform simple models as \textit{CV} and reach a very small median lateral prediction error of less than 0.21\,m at a prediction horizon of 5\,s. As shown in \autoref{tab:reference_Lat_Pred}, this is remarkable compared to other approaches. Note that we did not include studies in this compilation, which report the root-mean-square error (\textit{RMSE}), which we quantify with a value of 0.64\,m. On one hand, we follow \cite{willmott2005advantages}, which points out that \textit{RMSE} measures do not allow for a comparison over different data sets, as the values depend on the size of the data set. On the other, the challenge tackled by us (cf. \autoref{subsec:problem_statement}) is to predict the probability distribution of future vehicle positions rather than single shot estimates. Consequently, we did not optimize the predictions to minimize $RMSE$. Therefore, it is not surprising that other works which explicitly minimize this value, but ignore distribution estimations, perform better with respect to $RMSE$.

\begin{table}[!ht]
	\caption{Comparing Lateral Prediction Performance with Related Works}
	\label{tab:reference_Lat_Pred}
	\centering
	\begin{tabular}{|c|c|c|}
		\hline
	    &&\\[-0.8em]
		Approach & $\tilde{E}_{y,t}$ [m] & Prediction Horizon $t$ [s] \\
		\hline
		\cite{schlechtriemen2015will} & $\approx$ 0.18 & 5.0 \\
		\hline
		\cite{yoon2016} & 0.23 & 5.0 \\
		\hline
		\cite{wissing2018trajectory} & 0.50 & 3.0 \\
		\hline
		\textit{MLP} (PW-Raw) & 0.204 & 5.0\\
		\hline
	\end{tabular}
\end{table}


As shown in \cite{schlechtriemen2015will}, these results are dominated by the most frequent maneuver class ($FLW$). Hence, \autoref{tab:errors_per_maneuver}  complementarily shows the errors for 20\,000 maneuvers of each type.

\begin{table}[!ht]
	\caption{Prediction Errors per Class and Direction}
	\label{tab:errors_per_maneuver}
	\centering
	\begin{tabular}{|c|c|c|c|}
		\hline
		&&&\\[-0.8em]
		$d$ & $\tilde{E}_{d,5}^{LCL}$ [m] & $\tilde{E}_{d,5}^{FLW}$ [m]& $\tilde{E}_{d, 5}^{LCR}$ [m]\\		 
		 \hline
		 $x$ & 3.22 & 1.67 & 2.20 \\
		 \hline
		 $y$ & 1.25 & 0.19 & 1.80 \\
		\hline
	\end{tabular}
\end{table}

As can be seen, the errors for the lane change maneuvers are considerably larger than the ones for lane-following. On one hand, this can be explained with the more complex regression task. On the other, the predictions are subjected to higher uncertainties in case of a lane change, as shown by the predicted distributions (cf. \autoref{fig:predicted_distribution}). As opposed to that, the uncertainty is ignored in the single point estimates. Note that the increased uncertainties are caused by the lack of knowledge on the exact point in time at which the maneuver will be completed. This even holds true, if the classifier made the position prediction to know about an upcoming lane change.

\begin{figure}[!ht]
 \centering\includegraphics[width=0.95\columnwidth]{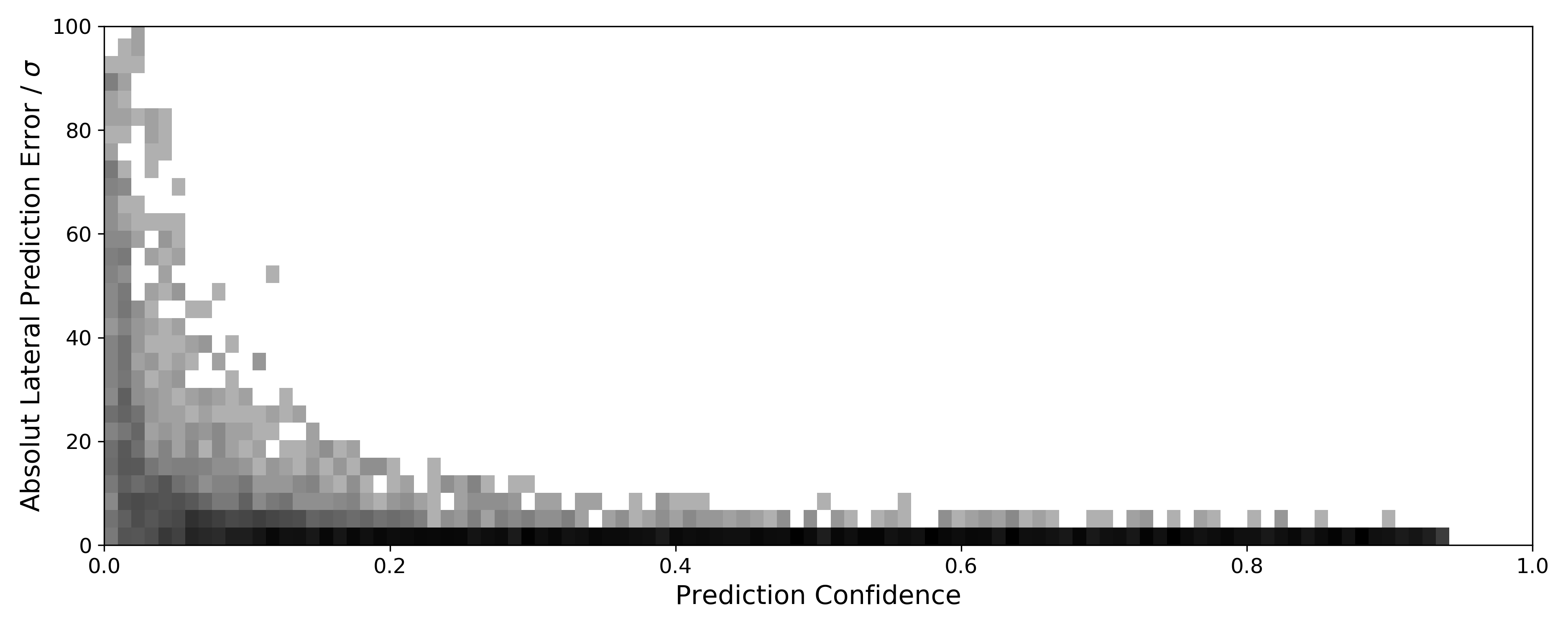}
 \caption{Prediction confidence against lateral prediction errors.}
 \label{fig:confidence}
\end{figure}

Complementary to these quantitative evaluations, we performed qualitative testing and visualized single situations along with our predictions. To illustrate this, we attached a short video and present a single frame in \autoref{fig:predicted_distribution}. More precisely, \autoref{fig:predicted_distribution} shows the predictions during an upcoming lane change, along with the described uncertainties. In addition, we show the confidence of our predictions ($Conf_x$, $Conf_y$), which provides an important hint concerning the reliability of the predictions to the consumer of the information. This value is calculated similarly to \cite{schlechtriemen2014probabilistic} through additional \textit{GMM}s fitted in the input dimensions. To demonstrate its general usability, we visualized the confidence value divided by the standard deviation against the lateral prediction errors at $T_h$=5\,s in \autoref{fig:confidence}. As can be seen, and as expected the prediction errors decrease with increasing confidence values.

\begin{table*}[!t]
	\renewcommand{\arraystretch}{1}
	\caption{Description of the Evaluated Features $f$ of an Observed Vehicle $o$ and Usage of the Features in the Constructed Feature Sets ($A$-$D$)}
	\label{tab:featureOverview}
	\centering
	\resizebox{2 \columnwidth}{!}{
	\begin{tabular}{|c|c|c|c|c|c|c|c|}
		\hline
		$\boldsymbol{R}$ & $\boldsymbol{f}$ &  \textbf{Description} & \textbf{Unit (Continuous)} & \multicolumn{4}{c|}{\textbf{Element of}} \\
		&  &  & \textbf{Range of Values (Nominal)} & \textbf{$B$} & \textbf{$C$} & \textbf{$D_{MLP}$} & $D_{GNB}$\\
		&  &  &  & $\vert \cdot \vert = 40$ & $\vert \cdot \vert= 29$ & $\vert \cdot \vert = 24$ & $\vert \cdot \vert = 48$\\
		\hline
		\hline
		& \multicolumn{7}{c|}{general information describing the related vehicle $r$ (cf. \autoref{eq:fsit} \& \autoref{fig:env_model})} \\
		\cline{2-8}
		$R_{rf}$&$actv^{r}$ & activity status & \{0: inactive, 1: active\}&\{f, l\}\footnotemark&\{f, fl, fr, l\}&\{fr, r\}&\{fl, fr, l, r\}\\
		\cline{2-8}
		&$mov^{r}$ & movement status & \{0: standing, 1: moving\}&\{f, l\}&\{f, fl, fr, l\}&\{r\}&\{fl, fr, r\}\\
		\cline{2-8}
		&\multirow{ 2}{*}{$class^{r}$} & \multirow{ 2}{*}{object class} &  \{0: bicycle, 1: motorbike, &\multirow{ 2}{*}{\{f, l\}}&\multirow{ 2}{*}{\{f, fl, l\}}&&\multirow{ 2}{*}{\{fl, fr, r\}}\\
		&&& 2: car, ..., 14: no class\}&&&&\\
		\cline{2-8}
		&\multirow{ 2}{*}{$cutinlvl^{r}$} & \multirow{ 2}{*}{cut-in level} & \{0: $P\leq0.5$, 1: $P>0.5$,&\multirow{ 2}{*}{\{l\}}&&&\multirow{ 2}{*}{\{r\}}\\
		&&& 2: $P > 0.66$, 3: $P > 0.9$\}&&&&\\
		\cline{2-8}
		& \multicolumn{7}{c|}{relation between $o$ and related vehicle $r$ in $o$'s coordinate-system} \\
		\cline{2-8}
		&$d^{rel, r}_{x}$ & longitudinal distance & \multirow{ 2}{*}{$m$} &\{f, l\}&\{f, l\}&\{f, r\}&\{fr, r\}\\
		\cline{2-3}\cline{5-8}
		&$d^{rel, r}_{y}$ & lateral distance &&\{f, l\}&\{f, fr\}&\{r\}&\{fr, r\}\\
		\cline{2-8}
		&$v^{rel, r}_{x}$ & relative longitudinal speed & \multirow{ 2}{*}{$\nicefrac{m}{s}$}&\{f, r\}&&\{r\}&\{f, fr, r\}\\
		\cline{2-3}\cline{5-8}
		&$v^{rel, r}_{y}$ & relative lateral speed &&\{f, fl, l, r\}&\{f\}&\{f, fr, r\}&\{f, fr\}\\
		\cline{2-8}
		&$a^{rel, r}_{x}$ & relative longitudinal acceleration &$\nicefrac{m}{s^2}$&&&\{fr\}&\{f, fl, fr, r\}\\
		\cline{2-8}
		& \multicolumn{6}{c|}{relation between $o$ and related vehicle $r$ in curvilinear coordinates} \\
		\cline{2-8}
		&$d^{rel, r}_{v}$ & longitudinal distance& \multirow{ 2}{*}{$m$}&\{f, l\}&\{l\}&\{fr\}&\{fl, fr, r\}\\
		\cline{2-3}\cline{5-8}
		&$d^{rel, r}_{u}$ & lateral distance&&\{f, l\}&\{fr\}&\{r\}&\{fr, r\}\\
		\cline{2-8}
		&$v^{rel, r}_{v} $ &relative longitudinal speed &  \multirow{ 2}{*}{$\nicefrac{m}{s}$}&\{f, r\}&&\{f\}&\{f, fl, fr, r\}\\
		\cline{2-3}\cline{5-8}
		&$v^{rel, r}_{u} $ &relative lateral speed & &\{f, fl, l, r\}&&\{l\}&\{fr, r\}\\
		\hline
		\hline
		$R_{rb}$&$mov^{r}$ & movement status & \{0: standing, 1: moving\}&\{rr\}&\{rr\}&\{rr\}&\{rr\}\\
		\cline{2-8}
		&$d^{rel, r}_{y}$  & lateral distance & $m$&\{rl\}&&\{rl\}&\{rl, rr\}\\
		\hline
		\hline
		$F_{o}$ &$ fog^{f}$ & status of the front fog lamp & \multirow{ 4}{*}{\{0: off, 1: on\}}&&&&\\
		\cline{2-3}\cline{5-8}
		&$ fog^{r}$ & status of the rear fog lamp & && &&\\ 
		\cline{2-3}\cline{5-8}
		&$ fog^{rl}$ & status of the rear left fog lamp && && &\\ 
		\cline{2-3}\cline{5-8}
		&$ fog^{rr}$ & status of the rear right fog lamp && && &\\
		\cline{2-8}
		&$ wpr$ & wiper level & \{0, ..., 15\} && &&\\ 
		\cline{2-8}
		& $d^{ml}_{y}$ & distance between the center of $o$ and the left marking & \multirow{ 4}{*}{$m$}&\cmark&\cmark&\cmark&\cmark\\
		\cline{2-3}\cline{5-8}
		&$ {d^{mr}_{y}}$ & distance between the center of $o$ and the right marking &&\cmark&\cmark&&\\
		\cline{2-3}\cline{5-8}
		&\multirow{ 2}{*}{$ d^{cl}_{y}$} & distance between the center of $o$ and & &\multirow{ 2}{*}{\cmark} &\multirow{ 2}{*}{\cmark}&&\\ 
		& & the centerline of the assigned lane && &&&\\ 
		\cline{2-8}
		&$v_x$ & longitudinal speed of the observed vehicle&\multirow{ 2}{*}{$\nicefrac{m}{s}$}&& &&\\
		\cline{2-3}\cline{5-8}
		&$v_y$ & lateral speed of the observed vehicle &&\cmark&\cmark&&\cmark\\
		\cline{2-8}
		&$a_x$ & longitudinal acceleration of the observed vehicle&\multirow{ 2}{*}{$\nicefrac{m}{s^2}$}&\cmark& &\cmark&\\
		\cline{2-3}\cline{5-8}
		&$a_y$ & lateral acceleration of the observed vehicle & & \cmark & \cmark & \cmark&\cmark\\
		\cline{2-8}
		&\multirow{ 2}{*}{$\psi$}  & angle of the observed vehicle & \multirow{ 2}{*}{$^\circ$} &\multirow{ 2}{*}{\cmark}&\multirow{ 2}{*}{\cmark}& \multirow{ 2}{*}{\cmark}& \multirow{ 2}{*}{\cmark} \\
		& & relative to the direction of the lane & &&&&\\
		\hline
		\hline
		$F_{infra}$ &$t^{ml}$ & type of the left marking & \{0: no marking, 1: continuous, &\cmark&\cmark&\cmark&\cmark\\
		\cline{2-3}\cline{5-8}
		&$t^{mr}$ & type of the right marking & 2: broken\}&\cmark&\cmark&\cmark&\cmark\\
		\cline{2-8}
		&$c^{ml}$ & color of the left marking & \{0: no marking, 1: white, && &&\cmark\\
		\cline{2-3}\cline{5-8}
		&$c^{mr}$ & color of the right marking & 2: yellow\}&& &&\cmark\\
		\cline{2-8}
		&$nlanes_{cam}$& number of parallel lanes observed via the camera& \{0: 0, ..., 3: 3+\}&&&&\cmark\\
		\cline{2-8}
		&$nlanes_{map}$& number of lanes stored in the map& \{0, ..., 5\}&& &&\\
		\cline{2-8}
		&$cntr$ & country & \{0: GER, 1: US, ...\}&&& &\\
		\cline{2-8}
		&$tnl$ & indicator if the situation takes place in a tunnel & \multirow{ 2}{*}{\{0: False, 1: True\}}&&&&\cmark\\
		\cline{2-3}\cline{5-8}
		&$brd$ &indicator if the situation takes place on a bridge & && &&\\
		\cline{2-8}
		&$v^{lim}$ & speedlimit of the current highway section & \{1: $>130{\frac{km}{h}}$, ..., 8: $<11{\frac{km}{h}}$\}&&& &\\
		\cline{2-8}
		&\multirow{ 2}{*}{$t^{a}$} & type of next approach & \{0: unknown, 1: on ramp, &\multirow{ 2}{*}{} & &&\\
		& & to the highway & 2: highway merge\}& && &\\
		\cline{2-8}
		&\multirow{ 2}{*}{$t^{e}$} & type of next exit & \{0: unknown, 1: ramp, &\multirow{ 2}{*}{}& &&\\
		& & of the highway & 2: highway divider\}&& &&\\
		\cline{2-8}
		&$w^{ml}$ & width of the left marking & \multirow{ 5}{*}{$m$}&\cmark& \cmark& &\\
		\cline{2-3}\cline{5-8}
		&$w^{mr}$ & width of the right marking &&\cmark& \cmark&&\\ 
		\cline{2-3}\cline{5-8}
		&$w^{lane}$ & width of the lane & && &\cmark&\\ 
		\cline{2-3}\cline{5-8}
		&$d_{x}^a$ & distance to the next approach to the highway & && &&\\ 
		\cline{2-3}\cline{5-8}
		&$d_{x}^e$ & distance to the next exit of the highway & && &&\\  
		\cline{2-8}
		& $c_{0}$ & curvature of the road & $\nicefrac{1}{m}$ && &&\\
		\cline{2-8}
		& $c_{1}$ & derrivation of the curvature & $\nicefrac{1}{m^2}$ && &&\\
		\hline
		\multicolumn{8}{l}{\footnotemark[\thefootnote] This means for example, that feature set $B$ (introduced in \autoref{subsec:feature_selection}), contains in total 40 elements, including the activity status of its surrounding vehicles in front (f)}\\
		\multicolumn{8}{l}{and to its left (l). In contrast to that, the activity states of its other front relation partners (r, fl, fr) are not included in $B$.}\\
	\end{tabular}
 }
 \vspace{-6mm}
\end{table*}

\section{Summary and Outlook}\label{sec:conclusion}
This work introduces a machine learning workflow that enables calculations of long-term behavior predictions for surrounding vehicles in highway scenarios. For the first time, a combined compilation of prediction techniques for driving maneuvers and positions as well as lateral and longitudinal behavior is presented. The developed modules are evaluated in detail based on a large amount of real-world data, challenging established state-of-the-art approaches.

To further improve the quality of the presented behavior predictions, especially in complex situations, we are working on various enhancements and conducting additional studies. Currently, we migrate the prediction strategies to an experimental vehicle to enable detailed investigations regarding run time as well as resource usage. Meanwhile, we are about to apply our models to predict movements of surrounding vehicles in contrast to ego-vehicle movements. Besides, we plan to apply our predictor to a publicly available data set as highD \cite{highDdataset} or NGSIM to improve comparability. In addition, we want to investigate up to which maximum prediction horizon (beyond 5\,s), the maneuver detection produces useful insights.

Moreover, we see high potential in identifying demanding scenarios and explicitly integrating contextual knowledge (e.\,g. weather, traffic, time of day or local specialties) into our models. First experiments towards this direction have proven, that  contextual properties can have a considerable impact on driving behavior.

\section*{Acknowledgment}

The authors would like to thank \textit{Mercedes-Benz AG Research and Development} for providing real-world measurement data, which enabled us to perform our experiments. Furthermore, we would like to thank the \textit{Institute of Databases and Information Systems} at \textit{Ulm University} as well as Prof.~Dr.~Klaus-Dieter~Kuhnert from the \textit{Institute of Realtime Learning Systems} at \textit{the University of Siegen} for supporting our studies.





\bibliographystyle{ieeetr}

\bibliography{bib}
%

\clearpage

\begin{IEEEbiography}[{\includegraphics[width=1in,height=1.25in,clip,keepaspectratio]{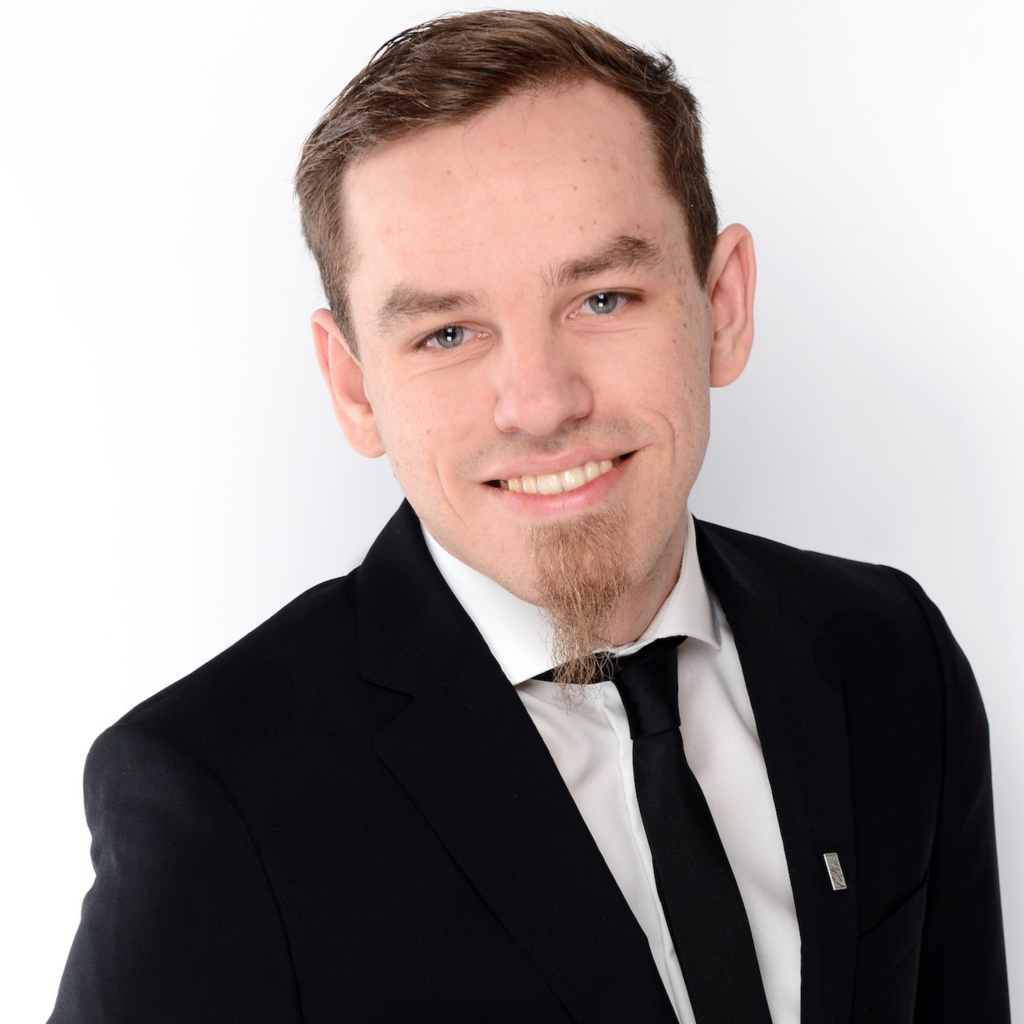}}]{Florian Wirthmüller}
received his B.\,Sc. and M.\,Sc. degrees in Computer Engineering with a study focus on Cognitive Technical Systems from Ilmenau University of Technology in 2015 and 2017, respectively. Since 2018 he has been working toward his Ph.\,D. degree with the Institute of Databases and Information Systems (DBIS) at Ulm University in cooperation with Mercedes-Benz AG Research and Development. His research interests include automated driving, big data analytics, machine learning, and backend architectures supporting manually \hspace*{1.1in} driven as well as automated vehicles.
\end{IEEEbiography}

\begin{IEEEbiography}[{\includegraphics[width=1in,height=1.25in,clip,keepaspectratio]{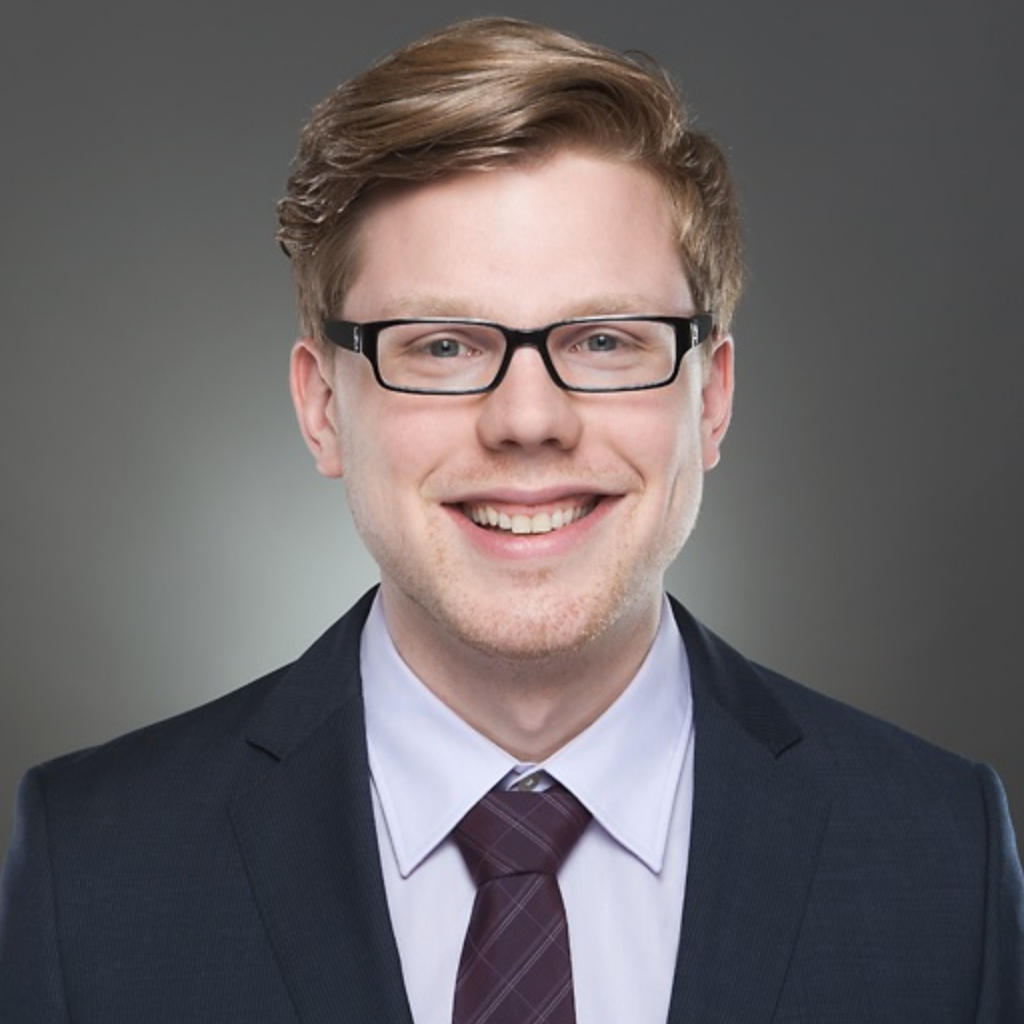}}]{Julian Schlechtriemen}
received his Diploma in Applied Computer Science with Electrical Engineering as main subject from the University of Siegen in 2012. Since 2012 he has been working toward his Ph.\,D. degree with the Institute of Realtime Learning Systems at the University of Siegen in cooperation with Mercedes-Benz AG Research and Development. His research interests include vehicle \& driver prediction using machine learning techniques and the incorporation of this information in behavior and trajectory planning. 
\end{IEEEbiography}

\begin{IEEEbiography}[{\includegraphics[width=1in,height=1.25in,clip,keepaspectratio]{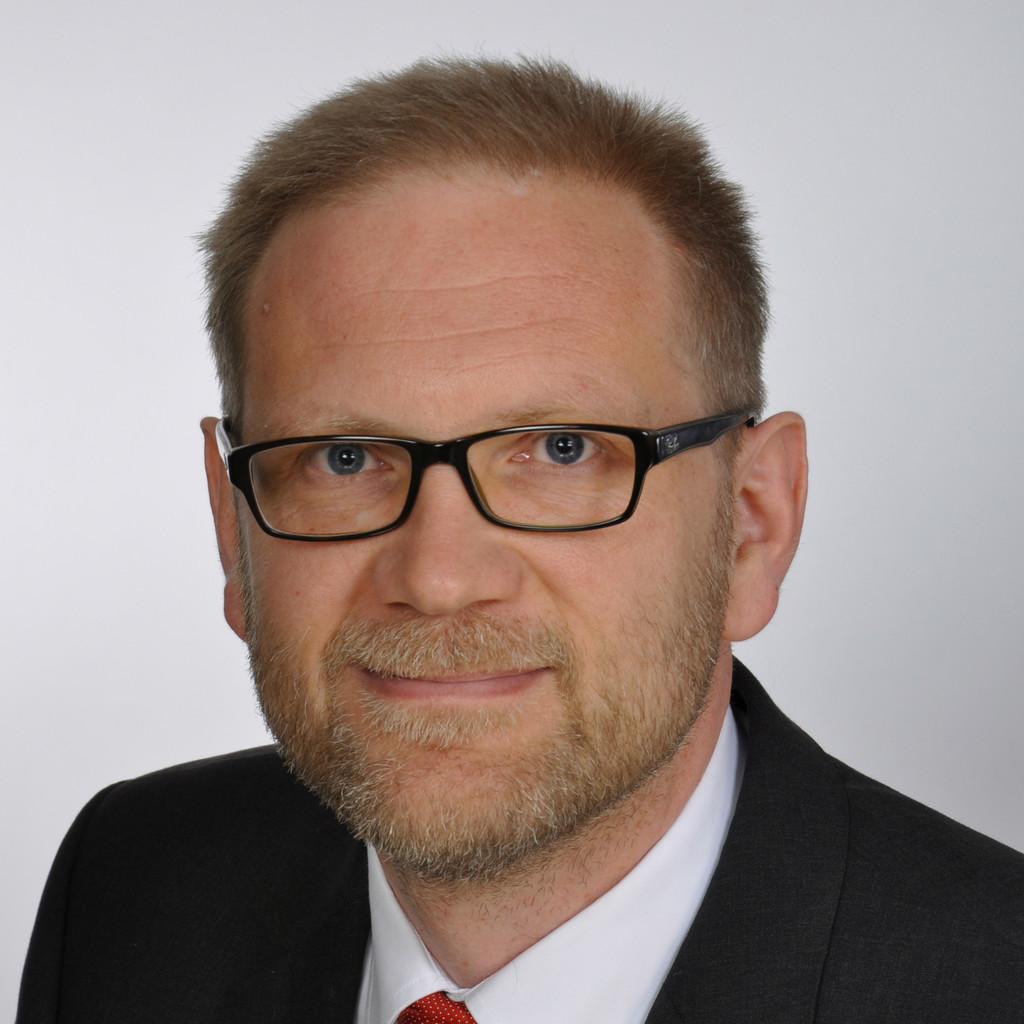}}]{Jochen Hipp} received his Diploma in Computer Science and Economics from the University of Tübingen. Since his Ph.\,D. deriving knowledge from massive data sets is part of his daily work at Mercedes-Benz AG Research and Development. Over the years he has been active in different fields like root cause analysis and early warning based on aftersales data, target-oriented endurance testing, customer profiles, advanced driver assistance systems, autonomous driving with a focus on high definiton maps, vehicle localization \hspace*{1.07in} and backend support. Today he is working on \hspace*{1.1in} the analysis of field data to improve current and \hspace*{1.1in} future driver assistance system generations.
\end{IEEEbiography}

\begin{IEEEbiography}[{\includegraphics[width=1in,height=1.25in,clip,keepaspectratio]{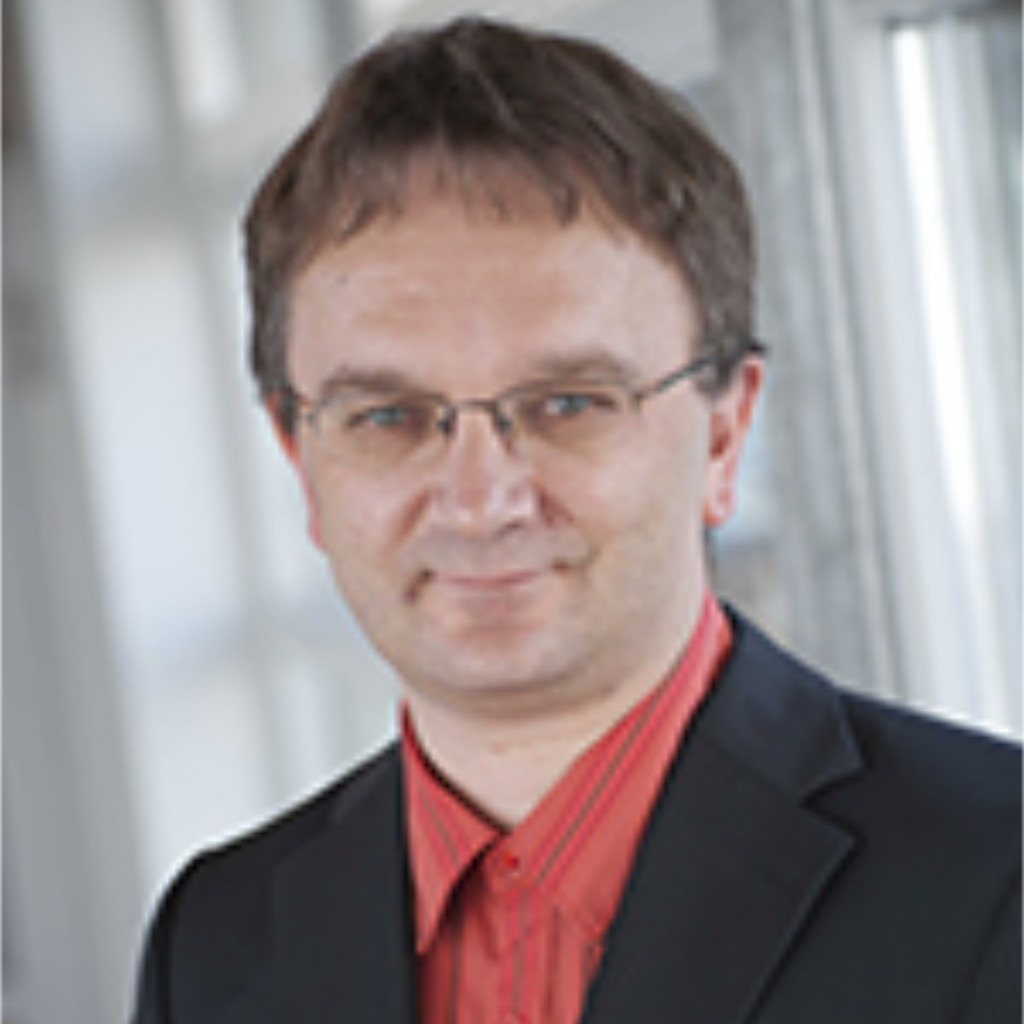}}]{Manfred Reichert} is a full professor at Ulm University where he is director of the Institute of Databases and Information Systems (DBIS). His research interests include business process management, intelligent information systems, process and data mining, and mobile services. He was PC Co-Chair of the BPM’08, CoopIS’11 and EDOC’13 conferences. Furthermore, he served as General Chair of the BPM’09 and EDOC’14 conferences as well as the BPM’15 workshops. Recently, he co-authored a Springer book on process flexibility and \hspace*{1.1in} obtained the BPM Test of Time Award at the BPM \hspace*{1.1in} 2013 conference.
\end{IEEEbiography}

\end{document}